\documentclass[final,5p,times,twocolumn]{elsarticle}

\usepackage{multirow}
\usepackage{amsmath}
\usepackage{graphicx}
\usepackage{natbib}
\usepackage{marvosym}
\usepackage{amssymb}
\usepackage{booktabs} 
\usepackage{color}
\usepackage{hyperref}
\usepackage{lineno}

\usepackage[linesnumbered,ruled]{algorithm2e}
\SetKwProg{Fn}{Function}{}{end}\SetKwFunction{FRecurs}{FnRecursive}%

\usepackage{etoolbox}

\journal{Swarm and Evolutionary Computation}

\begin{document}
\begin{frontmatter}



\title{Push and Pull Search Embedded in an M2M Framework for Solving Constrained Multi-objective Optimization Problems}







\author[addr1,addr2]{Zhun Fan\corref{cor1}}
\ead{zfan@stu.edu.cn}

\author[addr1,addr2]{Zhaojun Wang}
\author[addr1,addr2]{Wenji Li}
\author[addr1,addr2]{Yutong Yuan}
\author[addr1,addr2]{Yugen You}
\author[addr1,addr2]{Zhi Yang}
\author[addr1,addr2]{Fuzan Sun}
\author[addr1,addr2]{Jie Ruan}
\author[addr1,addr2]{Zhaocheng Li}

\cortext[cor1]{Corresponding author}

\address[addr1]{Department of Electronic Engineering, Shantou University, Guangdong, China}
\address[addr2]{Key Lab of Digital Signal and Image Processing of Guangdong Province, Shantou University, Guangdong, China}

\begin{abstract}

In dealing with constrained multi-objective optimization problems (CMOPs), a key issue of multi-objective evolutionary algorithms (MOEAs) is to balance the convergence and diversity of working populations. However, most state-of-the-art MOEAs show poor performance in balancing them, and can easily cause the working populations to concentrate on a part of regions of the Pareto fronts, leading to a serious imbalanced searching between preserving diversity and achieving convergence. This paper proposes a method which combines a multi-objective to multi-objective (M2M) approach with the push and pull search (PPS) framework, namely PPS-M2M. To be more specific, the proposed algorithm decomposes a CMOP into a set of simple CMOPs. Each simple CMOP corresponds to a sub-population and is solved in a collaborative manner. When dealing with constraints, each sub-population follows a procedure of "ignore the constraints in the push stage and consider the constraints in the pull stage", which helps each working sub-population get across infeasible regions. In order to evaluate the performance of the proposed PPS-M2M, it is compared with the other six algorithms, including M2M, MOEA/D-Epsilon, MOEA/D-SR, MOEA/D-CDP, C-MOEA/D and NSGA-II-CDP on a set of benchmark CMOPs. The experimental results show that the PPS-M2M is significantly better than the other six algorithms.
\end{abstract}


\end{frontmatter}


\section{Introduction}
Many real-world optimization problems can be formulated as constrained multi-objective optimization problems (CMOPs), which can be defined as follows \cite{kalyanmoy2001multi}:
\begin{equation}
\label{equ:cmop_definition}
\begin{cases}
\mbox{minimize} &\mathbf{F}(\mathbf{x})={(f_{1}(\mathbf{x}),\ldots,f_{m}(\mathbf{x}))} ^ {T} \\
\mbox{subject to} & g_i(\mathbf{x}) \ge 0, i = 1,\ldots,q \\
& h_j(\mathbf{x}) = 0, j= 1,\ldots,p \\
&\mathbf{x} \in {\mathbb{R}^n}
\end{cases}
\end{equation}
where $F(\mathbf{x})$ is an $m$-dimensional objective vector, and $F(\mathbf{x})\in \mathbb{R}^m$. ${g_i}(\mathbf{x}) \ge 0$ is an inequality constraint, and $q$ is the number of inequality constraints. ${h_j}(\mathbf{x})=0$ is an equality constraint, and $p$ represents the number of equality constraints. $\mathbf{x} \in \mathbb{R}^n$ is an $n$-dimensional decision vector.

In order to solve CMOPs with equality constraints, the equality constrains are often transformed into inequality constraints by using an extremely small positive number as follows:
\begin{eqnarray}
\label{equ:inequal2equal}
h_j(\mathbf{x})' \equiv \delta - |h_j(\mathbf{x})|  \ge 0
\end{eqnarray}

To deal with a set of constraints in CMOPs, an overall constraint violation is employed as follows:
\begin{eqnarray}
\label{equ:constraint}
\phi(\mathbf{x}) = \sum_{i=1}^{q} |\min(g_i(\mathbf{x}),0)| + \sum_{j = 1}^{p} |\min(h_j(\mathbf{x})',0)|
\end{eqnarray}

Given a solution $\mathbf{x}^{k} \in {\mathbb{R}^n}$, if $\phi(\mathbf{x}^{k}) = 0$, $\mathbf{x}^{k}$ is a feasible solution, otherwise it is infeasible. All feasible solutions form a feasible solution set.

To deal with constraints, Deb et al.\cite{996017} defined a $constraint$-$domination$ principle (CDP), that is, if a solution $x^{p}$ is said to constrained-dominate a solution $x^{q}$, one of the following conditions must be met:
\begin{enumerate}
\item Solution $x^{p}$ is a feasible solution and solution $x^{q}$ is an infeasible solution.
\item Both solution $x^{p}$ and solution $x^{q}$ are infeasible solutions, and the overall constraint violation of solution $x^{p}$ is smaller than that of solution $x^{q}$.
\item Both solution $x^{p}$ and solution $x^{q}$ are feasible solutions, and solution $x^{p}$ dominates solution $x^{q}$ in terms of objectives.
\end{enumerate}

A feasible solution set $FS=\{\phi(x)=0,x \in \mathbb{R}^n\}$ is constituted by all feasible solutions. Given a solution $\mathbf{x^{*}} \in FS $, if there is no any other solution $\mathbf{\overline{x}^{*}} \in FS$ satisfying $f_i(\mathbf{\overline{x}^{*}}) \leq f_i(\mathbf{x^{*}}) (i \in \{1,...,m\})$, $\mathbf{x^{*}}$ is called a Pareto optimal solution. All Pareto optimal solutions constitute a Pareto set (PS). The set of the mapping vectors of PS in the objective space is called a Pareto front (PF), which is defined as $PF = \{F(\mathbf{x})| \mathbf{x} \in PS\}$.

Over the past decade, a lot of research has been done in the field of multi-objective evolutionary algorithms (MOEAs) \cite{cai2018constrained} \cite{cai2018decomposition}. However, few works have been done in solving CMOPs. Recently, several constrained multi-objective evolutionary algorithms (CMOEAs) \cite{Fan2016Angle} \cite{fan2017improved} \cite{wang2019utilizing} \cite{liu2019handling} have been proposed to solve CMOPs. CMOEAs are particularly suitable for solving CMOPs, because they can find a number of Pareto optimal solutions in a single run, and are not affected by the mathematical properties of the objective functions. Therefore, the use of evolutionary algorithms to solve multi-objective optimization problems \cite{7927724} \cite{cai2015external} has become a research hot-spot in recent years.


To better balance minimizing the objectives and satisfying the constraints for CMOPs, many constraint-handling mechanisms have been designed, such as penalty function method \cite{5586543}, CDP \cite{deb2000efficient}, stochastic ranking \cite{runarsson2000stochastic}, $\varepsilon$-constrained \cite{takahama2010constrained}, and multi-objective concepts \cite{wang2007multiobjective}. For example, the penalty function methods transform constrained optimization problems into an unconstrained optimization problem by adding constraints multiplied by penalty factors to the objectives. If penalty factors remain constant throughout the optimization process for a period of time, it is called a static penalty method \cite{homaifar1994constrained}. If penalty factors are constructed as a function of the number of iterations or the iteration time, it is called a dynamic penalty approach \cite{joines1994use}. If penalty factors change according to the feedback information \cite{wang2017improving} during the search process, it is called an adaptive penalty approach \cite{woldesenbet2009constraint}.

Deb proposed a constraint-handling method called CDP \cite{deb2000efficient}, in which the fitness of a feasible solution is always better than that of an infeasible solution. Subsequently, CDP was extended to differential evolution (DE) by Mezura-Montes et al. \cite{mezura2006modified} to select target vectors and trial vectors. Moreover, CDP was used for designing parameter control in DE for constrained optimization \cite{mezura2009parameter}.

In order to overcome the weakness of penalty constraint-handling methods, stochastic ranking was proposed \cite{runarsson2000stochastic}, which uses a bubble-sort-like process to deal with the constrained optimization problems. Stochastic ranking uses a probability parameter $p_{f}\in[0,1]$ to determine whether the comparison is based on the objectives or based on the constraints. In the case of $p_{f}=0$, stochastic ranking is similar to the behavior of the feasibility rules. Furthermore, it can couple with various algorithms. For example, stochastic ranking has been combined with DE \cite{fan2009improved} and ant colony optimization \cite{leguizamon2007boundary}.

In this paper, we propose a new approach, namely PPS-M2M, to solve CMOPs, which combines a M2M approach \cite{Liu2014Decomposition} with push and pull search (PPS) framework \cite{fan2018push}. Unlike other constraint-handling mechanisms, the PPS-M2M decomposes a CMOP into a number of simple constrained multi-objective optimization sub-problems in the initial phase. Each sub-problem corresponds to a sup-population. During the search process, each sub-population evolves in a collaborative manner to ensure the diversity of the population. Inspired by the idea of information feedback model, some information about the constrained landscape is collected to help the parameters setting in the constraint-handling mechanisms.

In addition, the PPS-M2M divides the search process into two different phases. In the first phase, each sub-population approaches as close as possible to the unconstrained PF without considering any constraints. In the second phase, each sub-population is pulled back to the constrained PF using some constraint-handling approaches. The pull search stage is divided into two parts: (1) Only an improved epsilon constraint-handling mechanism is used \cite{fan2017improved} to optimize each subproblem for the first 90\% generations; (2) In the last 10\% of the generations, all sub-populations are merged into one population. Then the $\varepsilon$-dominance \cite{laumanns2002combining} and an improved epsilon constraint-handling mechanism \cite{fan2017improved} work together to evolve the population. In summary,  the proposed PPS-M2M has the following advantages.
\begin{enumerate}
\item At the beginning of the search, the method decomposes the population into a set of sub-populations, and each sub-population searches for a different multi-objective sub-problem in a coordinated manner. In other words, the PFs of all these subproblems constitute the PF of a CMOP. So the computational complexity is reduced by limiting the operator to focus on each subpopulation, and the convergence and diversity of the population are effectively ensured.
\item When dealing with constraints, the method follows a procedure of "ignore the constraints first and consider the constraints second", so that infeasible regions encountered a distance before the true PF present literally no extra barriers for the working population.
\item Since the landscape of constraints has been probed in the unconstrained push searching stage, this information can be employed to guide the parameter settings for mechanisms of constraint handling in the pull search stage.
\end{enumerate}

The remainder of this paper is organized as follows. Section \ref{sec:related_work} introduces the general idea of PPS framework and the M2M decomposition approach. Section \ref{sec:proposed-method} gives an instantiation of PPS-M2M. Section \ref{sec:exper} designs a set of experiments to compare the proposed PPS-M2M with the other six CMOEAs, including M2M \cite{Liu2014Decomposition}, MOEA/D-Epsilon \cite{Yang:2014vt}, MOEA/D-SR \cite{jan2013study}, MOEA/D-CDP \cite{jan2013study}, C-MOEA/D \cite{asafuddoula2012adaptive} and NSGA-II-CDP \cite{996017}. Finally, conclusions are drawn in section \ref{sec:conc}.

\section{Related Work}
\label{sec:related_work}

In this section, we introduce the general idea of PPS framework and the M2M population decomposition approach, which are used in the rest of this paper.

\subsection{The general idea of PPS framework}

The push and pull search (PPS) framework was introduced by Fan et al. \cite{fan2018push}. Unlike other constraint handling mechanisms, the search process of PPS is divided into two different stages: the push search and the pull search, and follows a procedure of "push first and pull second", by which the working population is pushed toward the unconstrained PF without considering any constraints in the push stage, and a constraint handling mechanism is used to pull the working population to the constrained PF in the pull stage.

In order to convert from the push search stage to the pull search stage, the following condition is applied
\begin{eqnarray}
\label{equ:r_rate}
r_{k} \equiv \max \{ rz_{k}, rn_{k} \} \le \epsilon
\end{eqnarray}
where $\epsilon$ is a threshold, which is defined by users. In Eq. \eqref{equ:r_rate}, we set $\epsilon = 1e-3$ in this work. During the last $l$ generations, $rz_{k}$ is the rate of change of the ideal point according to Eq. \eqref{equ:z_rate}, and $rn_{k}$ is the rate of change of the nadir point according to Eq. \eqref{equ:n_rate}.
\begin{eqnarray}
\label{equ:z_rate}
rz_k = \max_{i = 1,\ldots,m}\{  \frac{|z^k_i - z^{k-l}_i|}{ \max \{|z^{k-l}_i|,\Delta\}} \}
\end{eqnarray}
\begin{eqnarray}
\label{equ:n_rate}
rn_k = \max_{i = 1,\ldots,m}\{  \frac{|n^k_i - n^{k-l}_i|}{ \max \{|n^{k-l}_i|,\Delta\}} \}
\end{eqnarray}
where $z^k = (z^k_1,\ldots,z^k_m), n^k = (n^k_1,\ldots,n^k_m)$ are the ideal and nadir points in the $k$-th generation. $z^{k-l} = (z^{k-l}_1,\ldots,z^{k-l}_m)$, $n^{k - l} = (n^{k-l}_1,\ldots,n^{k-l}_m)$ are the ideal and nadir points in the $(k-l)$-th generation. $rz_k$ and $rn_k$ are two points in the interval $[0,1]$. $\Delta$ is a very small positive number, which is used to make sure that the denominators in Eq.\eqref{equ:z_rate} and Eq.\eqref{equ:n_rate} are not equal to zero. In this paper, $\Delta$ is set to $1e-6$. When $r_{k}$ is less than $\epsilon$, the push search stage is completed and the pull search stage is ready to start.

The major advantages of the PPS framework include:
\begin{enumerate}
\item In the push stage, the working population conveniently gets across infeasible regions without considering any constraints, which voids the impact of infeasible regions encountered a distance before the true PF.
\item Since the landscape of constraints has already been estimated in the push search stage, valuable information can be collected to guide the parameter settings in the pull search stage, which not only facilitates the parameter settings of the algorithm, but also enhances its adaptability when dealing with CMOPs.
\end{enumerate}

\subsection{The M2M population decomposition approach}
Employing a number of sub-populations to solve problems in a collaborative way \cite{rizk2018novel} is a widely used approach, which can help an algorithm balance its convergence and diversity. One of the most popular methods is the M2M population decomposition approach \cite{Liu2014Decomposition}, which decomposes a multi-objective optimization problems into a number of simple multi-objective optimization subproblems in the initialization, then solves these sub-problems simultaneously in a coordinated manner. For this purpose, $K$ unit vectors $v^{1},...,v^{K}$ in $\mathbb{R}_{+}^{m}$ are chosen in the first octant of the objective space. Then $\mathbb{R}_{+}^{m}$ is divided into $K$ subregions $\Omega_{1},...,\Omega_{K}$, where $\Omega_{k} (k=1,...,K)$ is
\begin{eqnarray}
\label{equ:M2M-angle}
\Omega_k = \{\mathbf{u} \in \mathbf{R}_{+}^{m} | \langle \mathbf{u}, \mathbf{v}^{k}  \rangle \leq \langle \mathbf{u}, \mathbf{v}^{j} \rangle ~for ~any ~j = 1,...,K  \}
\end{eqnarray}
where $\langle \mathbf{u}, \mathbf{v}^{j} \rangle$ is the acute angle between $\mathbf{u}$ and $\mathbf{v}^{j}$. Therefore, the population is decomposed into $K$ sub-populations, each sub-population searches for a different multi-objective subproblem. Subproblem $P_k$ is defined as:

\begin{eqnarray}
\label{equ:M2M-selection}
\begin{cases}
\mbox{minimize} &\mathbf{F}(x)=(f_{1}(x),...,f_{m}(x))\\
\mbox{subject to} &x \in \prod_{i=1}^{n}[a_{i},b_{i}]\\
&F(x) \in \Omega_{k}
\end{cases}
\end{eqnarray}

Altogether there are $K$ subproblems, and each subproblem is solved by employing a sub-population. Moreover, each sub-population has $S$ individuals. In order to keep $S$ individuals for each subproblem, some selection strategies are used. If sub-population $M_{k}$ has less than $S$ individuals, then $S-|M_{k}|$ individuals from $Q$ (the entire population) are randomly selected and added to $M_{k}$.

A major advantage of M2M is that it can effectively balance diversity and convergence at each generation by decomposing a multi-objective optimization problem into multiple simple multi-objective optimization problems.


\section{An Instantiation of PPS-M2M}
\label{sec:proposed-method}

This section describes the details of an instantiation, which combines the M2M approach with the PPS framework in a non-dominated sorting framework to solve CMOPs.

\subsection{The M2M approach}

In the initial stage, PPS-M2M uses a decomposition strategy to decompose a CMOP into a set of sub-problems that are solved in a collaborative manner with each sub-problem corresponding to a sub-population. For the sake of simplicity, we assume that all the objective functions $f_ {1}(x),...,f_{m}(x)$ of a CMOP are non-negative. Otherwise, the objective function $f_ {i}(x)$ is replaced by $f_{i}(x)-\bar{f}_{i}$, where $\bar{f}_{i}$ is the minimum value of the objective function $f_{i}(x)$ found so far.

\begin{figure}
	\begin{tabular}{c}
		\begin{minipage}[t]{\linewidth}
			\centering\includegraphics[height = 4.5cm]{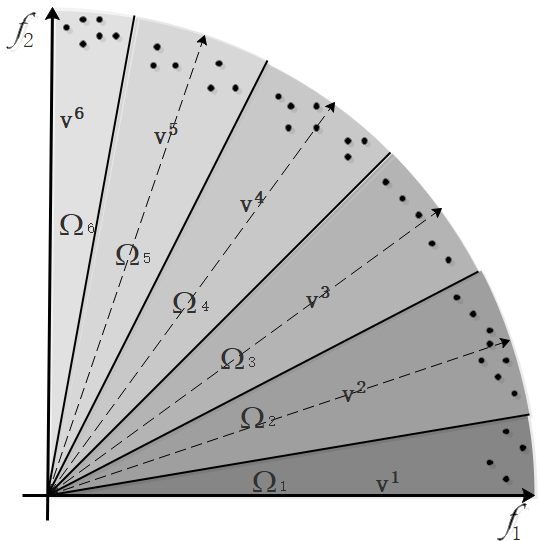}\\
		\end{minipage}
	\end{tabular}
\caption{\label{fig:M2M_fig} The population decomposition manner.}
\end{figure}

Assuming that the objective function space is divided into $K$ sub-regions, $K$ direction vectors $v^{1},...,v^{k}$ are uniformly distributed on the first octant of the unit sphere, where $v^{k}$ is the center of the $k$th subregion. Then the objective function space is divided into $K$ non-adjacent sub-regions $\Omega_{1},..., \Omega_{k}$, where the $k$th $(k=1,...,K)$ sub-regions can be obtained by the Eq.\eqref{equ:M2M-selection}. Through such a decomposition method, the multi-objective optimization problem (Eq. \eqref{equ:cmop_definition}) can be decomposed into $K$ simple CMOPs. The procedure is described in Algorithm \ref{alg:decomposition}. As an example when the objective number $m=2$ and the number of sub-regions $K=6$, the population decomposition is illustrated in Fig. \ref{fig:M2M_fig}, where $v_{1},...,v_{6}$ are six evenly distributed direction vectors.

\begin{algorithm}
 \Fn{result = AllocationSubPop($Q$,$K$)}{
    \For{$i \leftarrow 1$ \KwTo $K$}{
        Initialize $M_{k}$ as the solutions in $Q$ whose $F$-values are in $\Omega_k$ according to Eq. \eqref{equ:M2M-selection}; \\
        // $Q$: a set of individual solutions and their $F$-values.	
    }
    return $M_{1},...,M_{K}$;
 }
\caption{Allocation of Individuals to Sub-populations}
\label{alg:decomposition}
\end{algorithm}

\subsection{The PPS framework}

The search process of the PPS framework is divided into two main search stages: the push search stage and the pull search stage. In the push search stage, each sub-population uses an unconstrained NSGA-II to search for non-dominated solutions without considering any constraints. When using unconstrained NSGA-II to solve simple multi-objective optimization problems, an individual $i$ is measured by two attributes \cite{996017}, including the non-domination rank $i_{rank}$ and the crowding distance $i_{distance}$. Since individuals have two attributes, when an individual $i$ is compared to another individual $j$, if one of the following two requirements is satisfied, then individual $i$ can enter the descendant population.
\begin{enumerate}
\item If the individual $i$ dominates the individual $j$, i.e. $i_{rank}<j_{rank}$;
\item If the individual $i$ and the individual $j$ are non-dominated by each other, then the one with the larger crowding distance is selected, i.e. the condition $i_{rank}=j_{rank}$ is satisfied, and $i_{d}>j_{d}$.
\end{enumerate}
In other words, when two individuals belong to different non-domination rank, the individual with a lower (better) rank $i_{rank}$ is selected; when the two individuals have the same rank, the one with less crowding distance is selected.

In the push search stage, a constrained optimization problem is optimized without considering any constraints. The pseudocode of push search is given in Algorithm \ref{alg:push}. In line 2, non-dominated sorting is carried out on the sup-population $P_{k}$. In lines 3-6, a number of solutions are selected into $P_{k}^{'}$ until the number of solutions in $P_{k}^{'}$ is greater than $S_k$. Lines 7-9 select $S_k-|P_{k}^{'}|$ solutions into $P_{k}^{'}$ from $F_{i}$. In line 10, the sup-population $P_{k}$ is updated by setting $P_{k} = P_{k}^{'}$.
\begin{algorithm}
\Fn{PushSubproblems($P_{k}$,$S_k$)}{
    $F = $~nondominated-sort$(P_{k}),F=(F_{1},F_{2},...)$;\\
    $P_{k}^{'}=\varnothing ~and ~i=1$;\\
    $While (|P_{k}^{'}|+|F_{i}|\leq S_k)$\\
    $\qquad P_{k}^{'}=P_{k}^{'} \bigcup F_{i}$;\\
    $\qquad i=i+1$;\\
    calculate crowding-distance in $F_{i}$;\\
    sort solutions in $F_{i}$ by crowding-distance in a descending order \\
    $P_{k}^{'}=P_{k}^{'}\bigcup F_{i}[1:(S_k-|P_{k}^{'}|)]$;\\
    $P_{k} = P_{k}^{'}$;
}
\caption{Push Subproblems}
\label{alg:push}
\end{algorithm}

In the pull search stage, the constrained optimization problem is optimized by considering constraints, which is able to pull the population to the feasible and non-dominated regions. The pseudocode of push search is given in Algorithm \ref{alg:pull}.

The mechanism described in Eq. \eqref{equ:r_rate} controls the search process to switch from the push to the pull search. At the beginning of the evolutionary process, the value of $r_{k}$ is initialized to $1.0$ in order to ensure a thorough search in the push stage. The value of $r_{k}$ is updated by Eq.\eqref{equ:r_rate}. When the value of $r_{k}$ is less than or equal to the preset threshold $\epsilon$, the search behavior is changed.

In the pull stage, we need to prevent the population from falling into local optimum, and balance evolutionary search between feasible and infeasible regions. To achieve these goals, an improved epsilon constraint-handling mechanism \cite{fan2017improved} and the $\varepsilon$-dominance \cite{laumanns2002combining} are used to deal with constraints and objectives in the pull search stage, with the detailed formula given as follows.

An improved epsilon constraint-handling mechanism is defined as follows:
\begin{eqnarray}
\label{equ:IEpsilon}
& \epsilon(k)=
\begin{cases}
\phi_\theta\quad \quad \quad \quad \quad \text{if}\ \ k=0\\
(1 - \tau) \varepsilon(k-1)\  \text{if }rf_k < \alpha\\
\varepsilon(0)(1 - \frac{k}{T_c})^{cp}\ \  \text{if }rf_k \ge \alpha\\
0 \quad \quad \quad \quad \ \ \ \ \ \ \text{otherwise}
\end{cases}
\end{eqnarray}
where $\epsilon(k)$ is the value of $\epsilon$ function, $\phi_\theta$ is the overall constraint violation of the top $\theta$-th individual in the initial population, $rf_k$ is the proportion of feasible solutions in the generation $k$. $\tau$ controls the speed when the relaxation of constraints reduces in the case of $rf_k < \alpha$ ($\tau \in [0,1]$). $\alpha$ controls the searching preference between the feasible and infeasible regions. $cp$ is used to control the reducing interval of relaxation of constraints in the case of $rf_k \geq \alpha$. $\varepsilon(k)$ stops updating until the generation counter $k$ reaches generation $T_c$. $\varepsilon(0)$ is set to the maximum overall constraint violation when the push search finishes. In the case of $rf_k < \alpha$, Eq.\eqref{equ:IEpsilon} sets $\varepsilon(k)$ with an exponential decreasing speed, which has a potential to find feasible solutions more efficiently than the $\varepsilon$ setting in \cite{takahama2006constrained}. In the case of $rf_k \geq \alpha$, the Eq.\eqref{equ:IEpsilon} has the same $\varepsilon$ setting as adopted in \cite{takahama2006constrained}. In the pull search stage, a new individual is selected using the constraint handling mechanism described by Algorithm \ref{alg:pull}.
\begin{algorithm}
\Fn{result = PullSubproblems($M_{k}$,$gen$,$T_{max}$,$e$)}{
    //$T_{max}$: the maximum generation.\\
    //$e$: parameter setting in $\varepsilon$-domination;\\
    $result = false$;\\
    \eIf{$gen \le 0.9T_{max}$ }{
        \For{$k \leftarrow 1$ \KwTo $K$}{
           An improved epsilon constraint-handling mechanism is used to search for non-dominated and feasible solutions in $M_{k}$;
        }
    }{
        An improved epsilon constraint-handling mechanism and the $\varepsilon-dominance$ \cite{laumanns2002combining} are used to search non-dominated and feasible solutions in $M_{k}$;
    }
    \Return $result$;
}
\caption{Pull Subproblem}
\label{alg:pull}
\end{algorithm}

\subsection{PPS Embedded in M2M}


\begin{figure*}
	\begin{tabular}{cc}
		\begin{minipage}[t]{0.32\linewidth}  
			\includegraphics[width = 5cm]{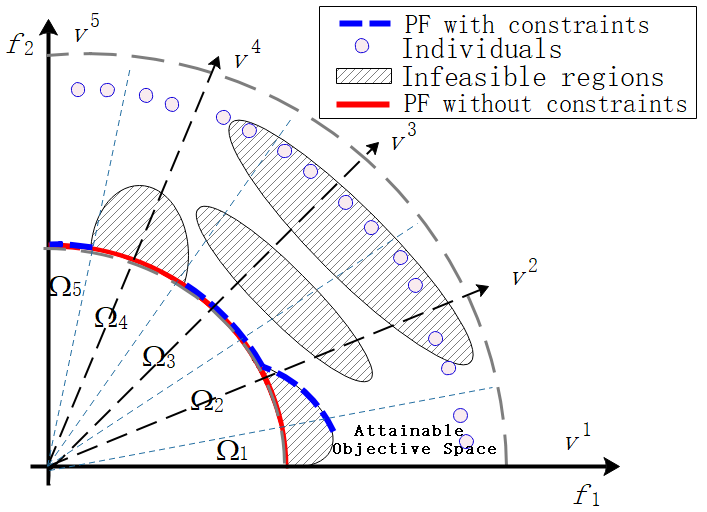}\\
			\centering{\scriptsize{(a)}}
		\end{minipage}
        \begin{minipage}[t]{0.32\linewidth}
			\includegraphics[width = 5cm]{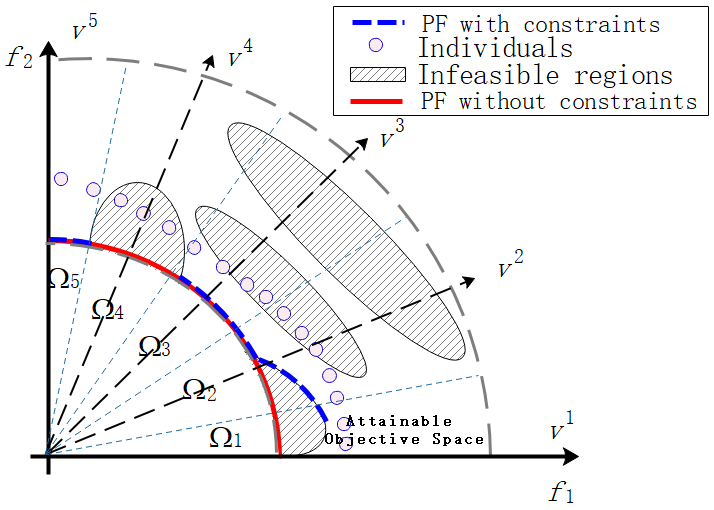}\\
			\centering{\scriptsize{(b)}}
		\end{minipage}
        \begin{minipage}[t]{0.32\linewidth}
			\includegraphics[width = 5cm]{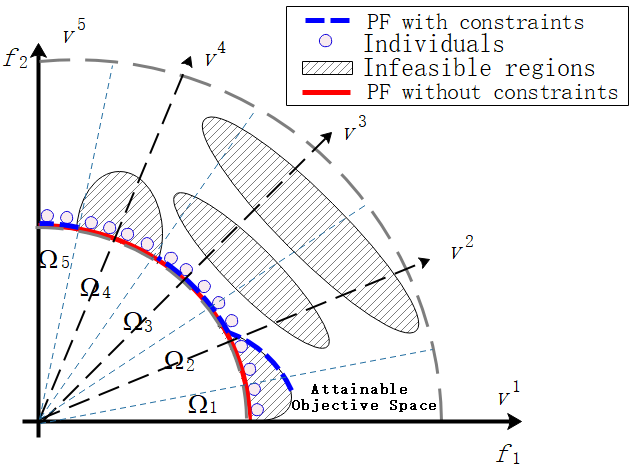}\\
			\centering{\scriptsize{(c)}}
		\end{minipage}
	\end{tabular}
	
    \begin{tabular}{cc}	
        \begin{minipage}[t]{0.32\linewidth}
			\includegraphics[width = 5cm]{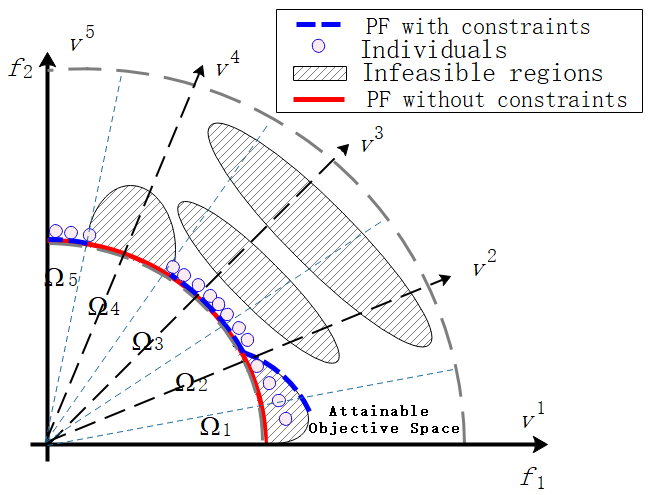}\\
			\centering{\scriptsize{(d)}}
		\end{minipage}
		\begin{minipage}[t]{0.32\linewidth}
			\includegraphics[width = 5cm]{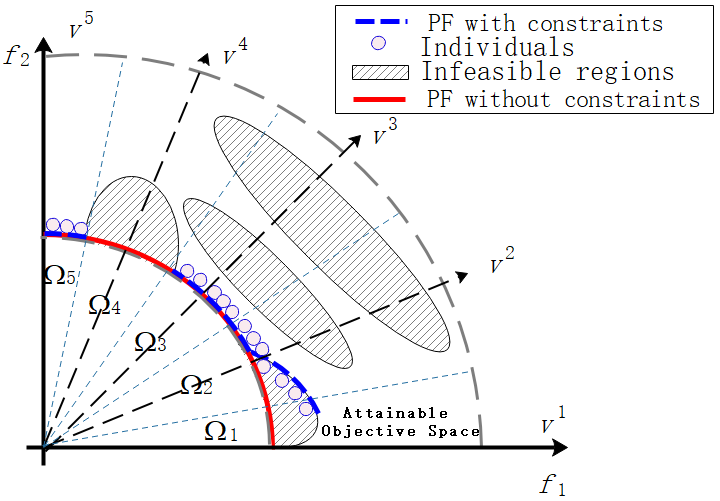}\\
			\centering{\scriptsize{(e)}}
		\end{minipage}
		\begin{minipage}[t]{0.32\linewidth}
			\includegraphics[width = 5cm]{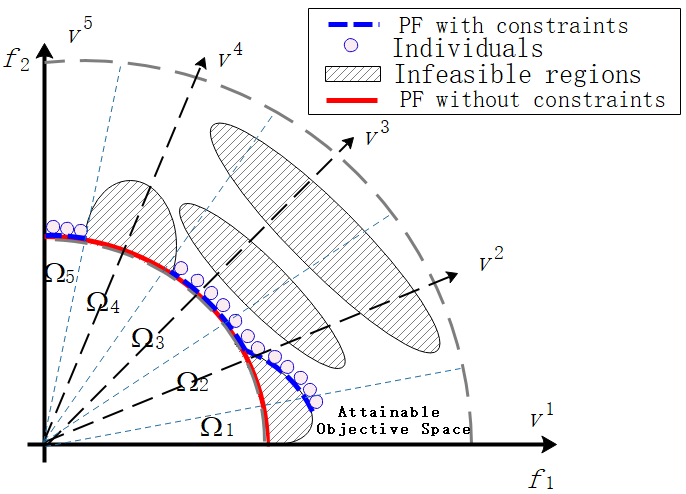}\\
			\centering{\scriptsize{(f)}}
		\end{minipage}
	\end{tabular}
	\caption{\label{fig:case} Infeasible regions make the original unconstrained PF partially feasible. The objective space is divided into 5 sub-regions, and 5 direction vector $v^{1},...,v^{5}$ are uniformly distributed on the first octant of the unit sphere. The objective space is divided into 5 non-adjacent sub-regions $M_{1},...,M_{5}$. (a)-(c) show the push search process, in which the working population of each sub-region crosses infeasible regions without any barriers. The pull search process, in which the infeasible solutions in the working population of each sub-region is pulled to the feasible and non-dominant regions, as shown in (d)-(f).}
\end{figure*}

The pseudo-code of PPS-M2M is introduced in Algorithm \ref{alg:PPS-M2M}. The algorithm is initialized at lines 1-3. At line 2, the population of a CMOP is divided into $K$ sub-populations, and the number of individuals for each sub-population is equal to $S$. At line 3, the maximum rate of change of ideal and nadir points $r_{k}$ is initialized to $1.0$, and the flag of search stage is set to push ($PushStage$=$true$). The algorithm runs repeatedly from line 4 to 38 until the termination condition is met. Lines 5-13 describe the process of generating new solutions for each sub-population. A number of new solutions are generated at lines 6-10. At lines 12-13, The solution set $Q$ is allocated to each sub-population according to Eq. \eqref{equ:M2M-angle}. The max rate of change between the ideal and nadir points during the last $l$ generations $r_{k}$ is calculated at line 15. The parameter $\varepsilon(k)$ is updated at lines 16-25. The updating process for each sub-population is described in lines 26-36. If the size of  sub-population $M_k$ is less than $S$, then $S-|M_k|$ individual solutions are randomly selected from $Q$ and added to $M_k$. If the size of sub-population $M_k$ is greater than $S$, then $S$ solutions are selected out by using the PPS framework. More specifically, at the push search stage, $S$ individual solutions are selected out by employing non-dominated sorting method without considering any constraints, as illustrated in line 31. At the pull search stage, $S$ individual solutions are selected out by using an improved epsilon constraint-handling approach, as illustrated in line 33. The generation counter is updated at line 37. At line 39, A set of non-dominated and feasible solutions is selected out.

As an example, the search behavior of PPS-M2M is illustrated in Fig. \ref{fig:case}, which can be summarized as follows. At first, five direction vectors $v^{1},...,v^{5}$ are uniformly sampled in the objective space, in which 5 non-adjacent sub-regions $M_{1},...,M_{5}$ are constructed. The working sub-populations have achieved the unconstrained PF without considering any constraints in the push search, as illustrated by Fig. \ref{fig:case}(a)-(c). It is notable that in this particular case some solutions located on the unconstrained PF are feasible. In the pull search stage, the infeasible solutions are pulled to the feasible and non-dominant regions, as illustrated by Fig. \ref{fig:case}(d)-(f).

\begin{algorithm}
    \KwIn{
    	\begin{enumerate}
    	\item[] $K$: the number of subproblems;
    	\item[] $K$ unit direction vectors: $v^{1}$,...,$v^{K}$;
    	\item[] $S$: the size of each subpopulation;
        \item[] $Q$: a set of individual solutions and their $F$-values;
    	\item[] $T_{c}$: the control generation for $\varepsilon(k)$;
        \item[] $e$: parameter setting in $\varepsilon$-domination;
        \item[] $T_{max}$: the maximum generation.
    	\end{enumerate}
    }
    \KwOut{a set of non-dominated and feasible solutions.}

	\textbf{Initialization:}\\
    Decompose a population into $K$ sub-populations ($M_{1},...,M_{K}$), each sub-population contains $S$ individuals according to $AllocationSubPop(Q,S,K)$;\\
    Set $r_{k}=1.0$, $PushStage$ = $true$;\\
    \While{$gen \le T_{max}$}{
    	
        \For{$k \leftarrow 1$ \KwTo $K$}{
            \ForEach{$x \in M_k$}{
               Randomly choose $y$ from $M_k$;\\
               Apply genetic operators on $x$ and $y$ to generate a new solution $z$;\\
               $R:=R \cup \{z\}$;\\
            }
           $Q:=R \cup (\cup_{k=1}^K M_k)$;\\
           Use $Q$ to set $M_1,...,M_K$ according to Eq. \eqref{equ:M2M-angle};\\
        }

    	  \lIf{$gen >= l$}{\\
          \qquad  Calculate $r_k$ according to Eq. \eqref{equ:r_rate}
    	 }
    	
    	 \eIf{$gen < T_c$}{
    	 	\eIf{$r_k \le \epsilon$ and $PushStage$ == $true$}{
    	 		\emph{$PushStage$ = false}; \\
    	 		$\varepsilon(gen)$ = $\varepsilon(0) =$ \emph{maxViolation};
    	 	}{
                Update $\varepsilon(gen)$ according to Eq. \eqref{equ:IEpsilon};
            }
    	 	
    	 }{
    	 	$\varepsilon(k)$ = 0;
    	 }
	
        \For{$k \leftarrow 1$ \KwTo $K$}{
            \eIf{$|M_k| \le S$ }{
                randomly select $S-|M_k|$ solutions from $Q$ and add them to $M_k$.\\
            }{
                \eIf{$PushStage$ == $true$}{
            	 	result = $PushSubproblems(M_k,S)$;
        	 	}{
                    result = $PullSubproblems(M_k,gen,T_{max},e)$;
                }
            }
        }

		$gen=gen+1$;\\
	}
Output the non-dominated and feasible solutions.

\caption{PPS-M2M}
\label{alg:PPS-M2M}
\end{algorithm}

\section{Experimental Study}
\label{sec:exper}

\subsection{Experimental Settings}
\label{sec:exper_setting}

The proposed PPS-M2M is compared with the other six algorithms(M2M \cite{Liu2014Decomposition}, MOEA/D-Epsilon \cite{Yang:2014vt}, MOEA/D-SR \cite{jan2013study}, MOEA/D-CDP \cite{jan2013study}, C-MOEA/D \cite{asafuddoula2012adaptive} and NSGA-II-CDP \cite{996017}) on LIR-CMOP1-14 \cite{fan2017improved}. The experimental parameters are listed as follows:

\begin{enumerate}
\item Population size: $N = 300$.
\item Halting condition: each algorithm runs for 30 times independently, and stops when 300,000 function evaluations are reached.
\item For the bi-objective instances, $K=10$. For the tri-objective instances, $K=15$.
\item The $K$ direction vectors are uniformly selected from the unit sphere in the first octant.
\item Parameter setting in $\varepsilon$-domination: $e = 0.01$.
\item Parameter setting in an improved epsilon constraint-handling
mechanism: $\alpha = 0.95$, $\tau = 0.1$.
\item Parameter setting in MOEA/D-IEpsilon: $T_c = 800$, $\alpha = 0.95$, $\tau = 0.1$, and $\theta = 0.05N$.
\item Parameter setting in MOEA/D-Epsilon: $T_c = 800$, $cp=2$, and $\theta = 0.05N$.
\item Parameter setting in MOEA/D-SR: $S_r = 0.05$.
\end{enumerate}

\subsection{Performance Metric}

In order to measure the performance of the proposed algorithm(PPS-M2M) and the other six algorithms(M2M \cite{Liu2014Decomposition}, MOEA/D-Epsilon \cite{Yang:2014vt}, MOEA/D-SR \cite{jan2013study}, MOEA/D-CDP \cite{jan2013study}, C-MOEA/D \cite{asafuddoula2012adaptive} and NSGA-II-CDP \cite{996017}), two performance indicators are used: the inverted generation distance (IGD) \cite{bosman2003balance} and the hypervolume\cite{Zitzler1999Multiobjective}.

\begin{itemize}
\item \textbf{Inverted Generational Distance} (IGD):
\end{itemize}

Inverted Generational Distance(IGD) is an inverse mapping of Generational Distance(GD). It is expressed by the average distance from the individual in Pareto optimal solution set to the non-dominant solution set $PF$ obtained by the algorithm. Therefore, the calculation formula is
\begin{equation} \label{IGD metric}
\begin{cases}
IGD(P^*,A) = \frac{\sum \limits_{y^* \in P^*}d(y^*,A)}{| P^* |}\\
\\
d(y^*,A) = \min \limits_{y \in A} \{ \sqrt {\sum_{i = 1} ^m (y^{*}_{i} - y_i)^2} \}
\end{cases}
\end{equation}
where $P^{*}$ is a set of representative solutions in the true PF. $d(y^{*},A)$ represents the minimum Euclidean distance from point $y^{*}_{i}$ on the Pareto optimal surface to individual $y_i$ in $P^{*}$. The smaller the IGD value, the better the performance of the algorithm.

\begin{itemize}
\item \textbf{Hypervolume} ($HV$):
\end{itemize}

$HV$ has become a popular evaluation index, which reflects the closeness of the set of non-dominated solutions achieved by a CMOEA to the true PF. The performance of CMOEA is evaluated by calculating the hypervolume of the space surrounded by the non-dominant solution set and the reference point. The calculation formula is as follows:
\begin{equation}
HV(S)=VOL(\bigcup \limits_{x\in S} [f_1(x),z_1^r]\times ...[f_m(x),z_m^r] ) \\
\end{equation}
where $VOL(\cdot)$ is the Lebesgue measure, $m$ denotes the number of objectives, $\mathbf{z}^r=(z_1^r,...,z_m^r)^T$ is a user-defined reference point in the objective space. The bigger the HV value, the better the performance of the algorithm. The reference point is placed at 1.2 times the distance to the nadir point of the true PF. A larger value of HV indicates better performance regarding diversity and/or convergence.

\subsection{Discussion of Experiments}

Table \ref{tab:igd} and Table \ref{tab:hv} show the IGD and HV values of the seven algorithms on LIR-CMOP1-14. According to the Friedman aligned test, PPS-M2M achieves the highest ranking among the seven CMOEAs. The $p$-values calculated by the statistics of the Friedman aligned test are close to zero, which reveals the significant differences among the seven algorithms.

Table \ref{tab:Friedman_igd} and Table \ref{tab:Friedman_HV} show adjusted $p$-values of IGD and HV values for the Friedman Aligned test, and PPS-M2M is the control method. To compare the statistical difference among PPS-M2M and the other six algorithms, we perform a series of post-hoc tests. Since each adjusted $p$ value in Table \ref{tab:Friedman_igd} and Table \ref{tab:Friedman_HV} is less than the preset significant level 0.05. To control the Family-Wise Error Rate (FWER), a set of post-hoc procedures, including the Holm procedure \cite{Sture1979A}, the Holland procedure \cite{holland1987improved}, the Finner procedure \cite{Finner1993On}, the Hochberg procedure \cite{hochberg1988sharper}, the Hommel procedure \cite{hommel1988stagewise}, the Rom procedure \cite{Rom1990A} and the Li procedure \cite{Li2008A}, are used according to \cite{Derrac2011A}. We can conclude that PPS-M2M is significantly better than the other six algorithms.

\begin{table*}[htbp]
	\centering 
	\caption{IGD results of PPS-M2M and the other six CMOEAs on LIR-CMOP1-14. To facilitate the display of this table, Epsilon, CDP, and SR in this table are short for MOEA/D-Epsilon, MOEA/D-CDP, and MOEA/D-SR respectively. Friedman test at a 0.05 significance level is performed between PPS-M2M and each of the other six CMOEAs.}
    \scalebox{0.9}[0.9]{
	\begin{tabular}{ccccccccc}	
    \hline	
	\multicolumn{2}{c}{\textbf{Test Instance}} & \textbf{PPS-M2M} & \textbf{M2M} & \textbf{Epsilon} & \textbf{CDP}   & \textbf{SR}   & \textbf{C-MOEA/D} & \textbf{NSGA-II-CDP} \\
	\hline	
	\multirow{2}[0]{*}{LIRCMOP1} & mean  & 2.341E-02 & 3.106E-02 & 5.74E-02 & 1.11E-01 & \textbf{1.81E-02} & 1.26E-01 & 3.23E-01 \\
          & std   & 8.369E-03 & 9.888E-03 & 2.89E-02 & 5.04E-02 & 1.66E-02 & 7.03E-02 & 7.33E-02 \\
    \hline
    \multirow{2}[0]{*}{LIRCMOP2} & mean  & \textbf{1.604E-02} & 2.742E-02 & 5.39E-02 & 1.43E-01 & 9.63E-03 & 1.40E-01 & 3.03E-01 \\
          & std   & 7.606E-03 & 8.596E-03 & 2.13E-02 & 5.55E-02 & 7.23E-03 & 5.44E-02 & 7.24E-02 \\
    \hline
    \multirow{2}[0]{*}{LIRCMOP3} & mean  & \textbf{3.330E-02} & 4.383E-02 & 8.81E-02 & 2.61E-01 & 1.78E-01 & 2.80E-01 & 4.08E-01 \\
          & std   & 1.274E-02 & 1.445E-02 & 4.36E-02 & 4.33E-02 & 7.20E-02 & 4.21E-02 & 1.15E-01 \\
    \hline
    \multirow{2}[0]{*}{LIRCMOP4} & mean  & \textbf{3.738E-02} & 4.187E-02 & 6.51E-02 & 2.53E-01 & 1.95E-01 & 2.59E-01 & 3.85E-01 \\
          & std   & 1.421E-02 & 2.371E-02 & 3.01E-02 & 4.34E-02 & 6.40E-02 & 3.51E-02 & 1.35E-01 \\
    \hline
    \multirow{2}[0]{*}{LIRCMOP5} & mean  & \textbf{8.343E-03} & 2.941E-01 & 1.15E+00 & 1.05E+00 & 1.04E+00 & 1.10E+00 & 5.53E-01 \\
          & std   & 1.750E-03 & 5.013E-01 & 1.98E-01 & 3.63E-01 & 3.66E-01 & 2.99E-01 & 6.88E-01 \\
    \hline
    \multirow{2}[0]{*}{LIRCMOP6} & mean  & \textbf{9.631E-03} & 5.356E-01 & 1.27E+00 & 1.09E+00 & 9.43E-01 & 1.31E+00 & 5.74E-01 \\
          & std   & 1.711E-03 & 5.509E-01 & 2.95E-01 & 5.20E-01 & 5.90E-01 & 2.08E-01 & 4.21E-01 \\
    \hline
    \multirow{2}[0]{*}{LIRCMOP7} & mean  & \textbf{9.335E-03} & 5.237E-01 & 1.51E+00 & 1.46E+00 & 1.08E+00 & 1.56E+00 & 2.38E-01 \\
          & std   & 2.459E-03 & 7.677E-01 & 5.09E-01 & 5.58E-01 & 7.58E-01 & 4.24E-01 & 4.06E-01 \\
    \hline
    \multirow{2}[0]{*}{LIRCMOP8} & mean  & \textbf{9.351E-03} & 7.924E-01 & 1.62E+00 & 1.38E+00 & 1.01E+00 & 1.58E+00 & 6.02E-01 \\
          & std   & 2.143E-03 & 7.455E-01 & 3.05E-01 & 6.15E-01 & 7.24E-01 & 3.71E-01 & 7.39E-01 \\
    \hline
    \multirow{2}[0]{*}{LIRCMOP9} & mean  & \textbf{2.886E-01} & 4.599E-01 & 4.90E-01 & 4.81E-01 & 4.85E-01 & 4.81E-01 & 6.44E-01 \\
          & std   & 1.239E-01 & 6.426E-02 & 4.22E-02 & 5.24E-02 & 4.78E-02 & 5.24E-02 & 1.60E-02 \\
    \hline
    \multirow{2}[0]{*}{LIRCMOP10} & mean  & \textbf{1.894E-02} & 2.291E-01 & 2.13E-01 & 2.16E-01 & 1.92E-01 & 2.13E-01 & 5.97E-01 \\
          & std   & 5.322E-02 & 8.123E-02 & 5.32E-02 & 6.81E-02 & 6.81E-02 & 4.63E-02 & 3.20E-02 \\
    \hline
    \multirow{2}[0]{*}{LIRCMOP11} & mean  & \textbf{1.194E-02} & 4.400E-01 & 3.47E-01 & 3.42E-01 & 3.16E-01 & 3.81E-01 & 4.87E-01 \\
          & std   & 3.574E-02 & 1.038E-01 & 9.28E-02 & 9.22E-02 & 7.49E-02 & 8.95E-02 & 1.05E-02 \\
    \hline
    \multirow{2}[0]{*}{LIRCMOP12} & mean  & \textbf{8.071E-02} & 1.488E-01 & 2.52E-01 & 2.69E-01 & 2.06E-01 & 2.50E+00 & 5.80E-01 \\
          & std   & 6.031E-02 & 5.757E-02 & 8.98E-02 & 9.06E-02 & 5.61E-02 & 9.63E-02 & 1.17E-01 \\
    \hline
    \multirow{2}[0]{*}{LIRCMOP13} & mean  & \textbf{1.858E-01} & 9.751E-01 & 1.20E+00 & 1.21E+00 & 8.86E-01 & 1.18E+00 & 1.39E+01 \\
          & std   & 3.009E-02 & 5.292E-01 & 3.06E-01 & 3.17E-01 & 5.76E-01 & 3.78E-01 & 2.26E+00 \\
    \hline
    \multirow{2}[0]{*}{LIRCMOP14} & mean  & \textbf{1.759E-01} & 9.429E-01 & 1.02E+00 & 1.11E+00 & 1.03E+00 & 1.25E+00 & 1.36E+01 \\
          & std   & 3.257E-02 & 5.152E-01 & 4.86E-01 & 3.98E-01 & 4.70E-01 & 5.30E-02 & 2.17E+00 \\
    \hline
    \multicolumn{2}{c}{\textbf{Friedman Test}} & \textbf{1.1429} & 2.9286 & 4.6786 & 4.8929 & 3.1429 & 5.5714 & 5.6429 \\

	\bottomrule
	\end{tabular}}%
	\label{tab:igd}%
\end{table*}%

\begin{table*}[!htp]
	\centering \normalsize
	\caption{Adjusted p-values for the Friedman Aligned test in terms of mean metric (IGD).}

  \scalebox{0.9}[0.9]{
	\begin{tabular}{cccccccccc}
		\hline
		i&algorithm&unadjusted $p$&$p_{Holm}$&$p_{Hochberg}$&$p_{Hommel}$&$p_{Holland}$&$p_{Rom}$&$p_{Finner}$&$p_{Li}$\\
		\hline
		1 &NSGA-II-CDP &0&0&0&0&0&0&0&0 \\
        2 &C-MOEA/D &0&0&0&0&0&0&0&0 \\
        3 &CDP &0.000004&0.000017&0.000017&0.000017&0.000017&0.000017&0.000009&0.000005\\
        4 &Epsilon &0.000015&0.000045&0.000045&0.000045&0.000045&0.000045&0.000022&0.000015 \\
        5 &SR &0.014306&0.028612&0.028612&0.028612&0.028407&0.028612&0.017142&0.014515 \\
        6 &M2M &0.028739&0.028739&0.028739&0.028739&0.028739&0.028739&0.028739&0.028739 \\
		\hline
	\end{tabular}
  }
	\label{tab:Friedman_igd}
\end{table*}

\begin{table*}[htbp]
	\centering \normalsize
	\caption{HV results of PPS-M2M and the other six CMOEAs on LIR-CMOP1-14. To facilitate the display of this table, Epsilon, CDP, and SR in this table are short for MOEA/D-Epsilon, MOEA/D-CDP, and MOEA/D-SR respectively. Friedman test at a 0.05 significance level is performed between PPS-M2M and each of the other six CMOEAs.}
    \scalebox{0.9}[0.9]{
	\begin{tabular}{ccccccccc}
		\hline	
		\multicolumn{2}{c}{\textbf{Test Instance}} & \textbf{PPS-M2M} & \textbf{M2M} & \textbf{Epsilon} & \textbf{CDP}   & \textbf{SR}   & \textbf{C-MOEA/D} & \textbf{NSGA-II-CDP} \\
		\hline
		\multirow{2}[0]{*}{LIRCMOP1} & mean  & \textbf{1.003E+00} & 9.897E-01 & 9.590E-01 & 7.540E-01 & 9.960E-01 & 7.410E-01 & 5.160E-01 \\
          & std   & 7.312E-03 & 1.322E-02 & 3.280E-02 & 8.950E-02 & 2.910E-02 & 1.220E-01 & 5.570E-02 \\
        \hline
        \multirow{2}[0]{*}{LIRCMOP2} & mean  & 1.334E+00 & 1.321E+00 & 1.280E+00 & 1.060E+00 & \textbf{1.340E+00} & 1.070E+00 & 8.240E-01 \\
          & std   & 1.083E-02 & 1.124E-02 & 2.880E-02 & 1.080E-01 & 1.470E-02 & 9.100E-02 & 1.150E-01 \\
        \hline
        \multirow{2}[0]{*}{LIRCMOP3} & mean  & \textbf{8.499E-01} & 8.407E-01 & 7.980E-01 & 4.860E-01 & 5.910E-01 & 4.710E-01 & 4.080E-01 \\
          & std   & 1.423E-02 & 1.787E-02 & 3.930E-02 & 4.310E-02 & 1.070E-01 & 4.090E-02 & 2.880E-02 \\
        \hline
        \multirow{2}[0]{*}{LIRCMOP4} & mean  & \textbf{1.059E+00} & 1.051E+00 & 1.020E+00 & 7.350E-01 & 8.150E-01 & 7.310E-01 & 6.170E-01 \\
          & std   & 1.831E-02 & 3.870E-02 & 4.190E-02 & 5.440E-02 & 8.700E-02 & 5.160E-02 & 1.060E-01 \\
        \hline
        \multirow{2}[0]{*}{LIRCMOP5} & mean  & \textbf{1.451E+00} & 1.084E+00 & 4.300E-02 & 1.630E-01 & 1.820E-01 & 9.720E-02 & 9.390E-01 \\
          & std   & 4.778E-03 & 6.097E-01 & 2.350E-01 & 4.430E-01 & 4.390E-01 & 3.700E-01 & 3.210E-01 \\
        \hline
        \multirow{2}[0]{*}{LIRCMOP6} & mean  & \textbf{1.119E+00} & 5.337E-01 & 5.400E-02 & 1.880E-01 & 3.020E-01 & 2.330E-02 & 4.130E-01 \\
          & std   & 2.051E-03 & 3.987E-01 & 2.210E-01 & 3.870E-01 & 4.620E-01 & 1.280E-01 & 1.890E-01 \\
        \hline
        \multirow{2}[0]{*}{LIRCMOP7} & mean  & \textbf{3.002E+00} & 2.021E+00 & 3.030E-01 & 3.740E-01 & 9.880E-01 & 2.040E-01 & 2.400E+00 \\
          & std   & 8.630E-03 & 1.349E+00 & 9.070E-01 & 9.580E-01 & 1.270E+00 & 7.520E-01 & 6.520E-01 \\
        \hline
        \multirow{2}[0]{*}{LIRCMOP8} & mean  & \textbf{3.002E+00} & 1.498E+00 & 1.060E-01 & 5.170E-01 & 1.100E+00 & 1.660E-01 & 1.900E+00 \\
          & std   & 5.415E-03 & 1.277E+00 & 5.490E-01 & 1.050E+00 & 1.200E+00 & 6.110E-01 & 7.560E-01 \\
        \hline
        \multirow{2}[0]{*}{LIRCMOP9} & mean  & \textbf{3.287E+00} & 2.826E+00 & 2.740E+00 & 2.770E+00 & 2.750E+00 & 2.770E+00 & 2.060E+00 \\
          & std   & 1.926E-01 & 2.114E-01 & 1.480E-01 & 1.840E-01 & 1.640E-01 & 1.840E-01 & 1.080E-02 \\
        \hline
        \multirow{2}[0]{*}{LIRCMOP10} & mean  & \textbf{3.212E+00} & 2.845E+00 & 2.890E+00 & 2.880E+00 & 2.930E+00 & 2.890E+00 & 2.040E+00 \\
          & std   & 8.954E-02 & 1.720E-01 & 1.020E-01 & 1.360E-01 & 1.350E-01 & 9.770E-02 & 4.450E-02 \\
        \hline
        \multirow{2}[0]{*}{LIRCMOP11} & mean  & \textbf{4.359E+00} & 3.115E+00 & 3.340E+00 & 3.350E+00 & 3.380E+00 & 3.240E+00 & 3.110E+00 \\
          & std   & 1.167E-01 & 3.003E-01 & 2.570E-01 & 2.570E-01 & 2.900E-01 & 2.550E-01 & 1.540E-02 \\
        \hline
        \multirow{2}[0]{*}{LIRCMOP12} & mean  & \textbf{5.449E+00} & 5.221E+00 & 4.880E+00 & 4.830E+00 & 5.030E+00 & 4.890E+00 & 3.280E+00 \\
          & std   & 1.852E-01 & 1.878E-01 & 3.170E-01 & 3.280E-01 & 1.750E-01 & 3.450E-01 & 3.610E-01 \\
        \hline
        \multirow{2}[0]{*}{LIRCMOP13} & mean  & \textbf{4.856E+00} & 1.518E+00 & 4.550E-01 & 4.630E-01 & 1.890E+00 & 6.290E-01 & 0.000E+00 \\
          & std   & 1.378E-01 & 2.084E+00 & 1.300E+00 & 1.420E+00 & 2.570E+00 & 1.710E+00 & 0.000E+00 \\
        \hline
        \multirow{2}[0]{*}{LIRCMOP14} & mean  & \textbf{5.425E+00} & 1.744E+00 & 1.330E+00 & 8.810E-01 & 1.270E+00 & 1.800E-01 & 0.000E+00 \\
          & std   & 1.240E-01 & 2.314E+00 & 2.450E+00 & 1.970E+00 & 2.290E+00 & 2.600E-01 & 0.000E+00 \\
        \hline
        \multicolumn{2}{c}{\textbf{Friedman Test}} & \textbf{1.0714} & 2.9286 & 4.8214 & 4.8929 & 3.2143 & 5.3571 & 5.7143 \\
	\bottomrule
	\end{tabular}}%
	\label{tab:hv}%
\end{table*}%

\begin{table*}[!htp]
	\centering \normalsize
	\caption{Adjusted p-values for the Friedman Aligned test in terms of mean metric (HV)}
    \scalebox{0.9}[0.9]{
	\begin{tabular}{cccccccccc}
		\hline
		i&algorithm&unadjusted $p$&$p_{Holm}$&$p_{Hochberg}$&$p_{Hommel}$&$p_{Holland}$&$p_{Rom}$&$p_{Finner}$&$p_{Li}$\\
		\hline
        1 &NSGA-II-CDP &0  &0 &0 &0 &0 &0 &0 &0\\
        2 &C-MOEA/D &0&0.000001&0.000001&0.000001&0.000001&0.000001&0&0\\
        3 &CDP &0.000003&0.000011&0.000011&0.000009&0.000011&0.000011&0.000006&0.000003\\
        4 &Epsilon &0.000004&0.000013&0.000013&0.000013&0.000013&0.000013&0.000007&0.000004\\
        5 &SR &0.008679&0.017358&0.017358&0.017358&0.017282&0.017358&0.010406&0.008804\\
        6 &M2M &0.022934&0.022934&0.022934&0.022934&0.022934&0.022934&0.022934&0.022934\\
		\hline
	\end{tabular}
    }
	\label{tab:Friedman_HV}
\end{table*}

\begin{figure*}
	\begin{tabular}{cc}
		\begin{minipage}[t]{0.28\linewidth}
			\includegraphics[height = 4.5cm]{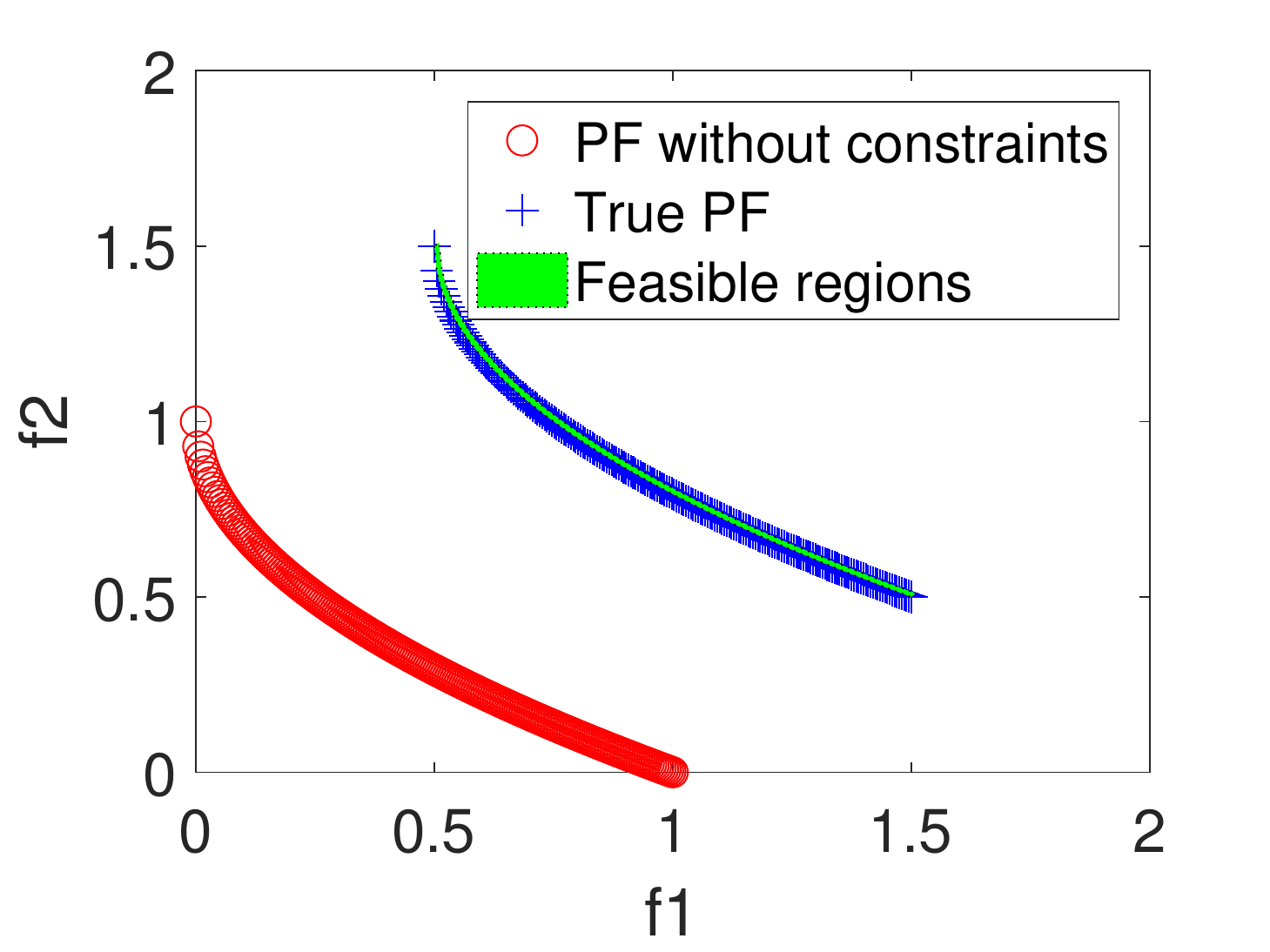}\\
			\centering{\scriptsize{(a) LIR-CMOP2}}
		\end{minipage}
		\hspace{0.5cm}
		\begin{minipage}[t]{0.28\linewidth}
			\includegraphics[height = 4.5cm]{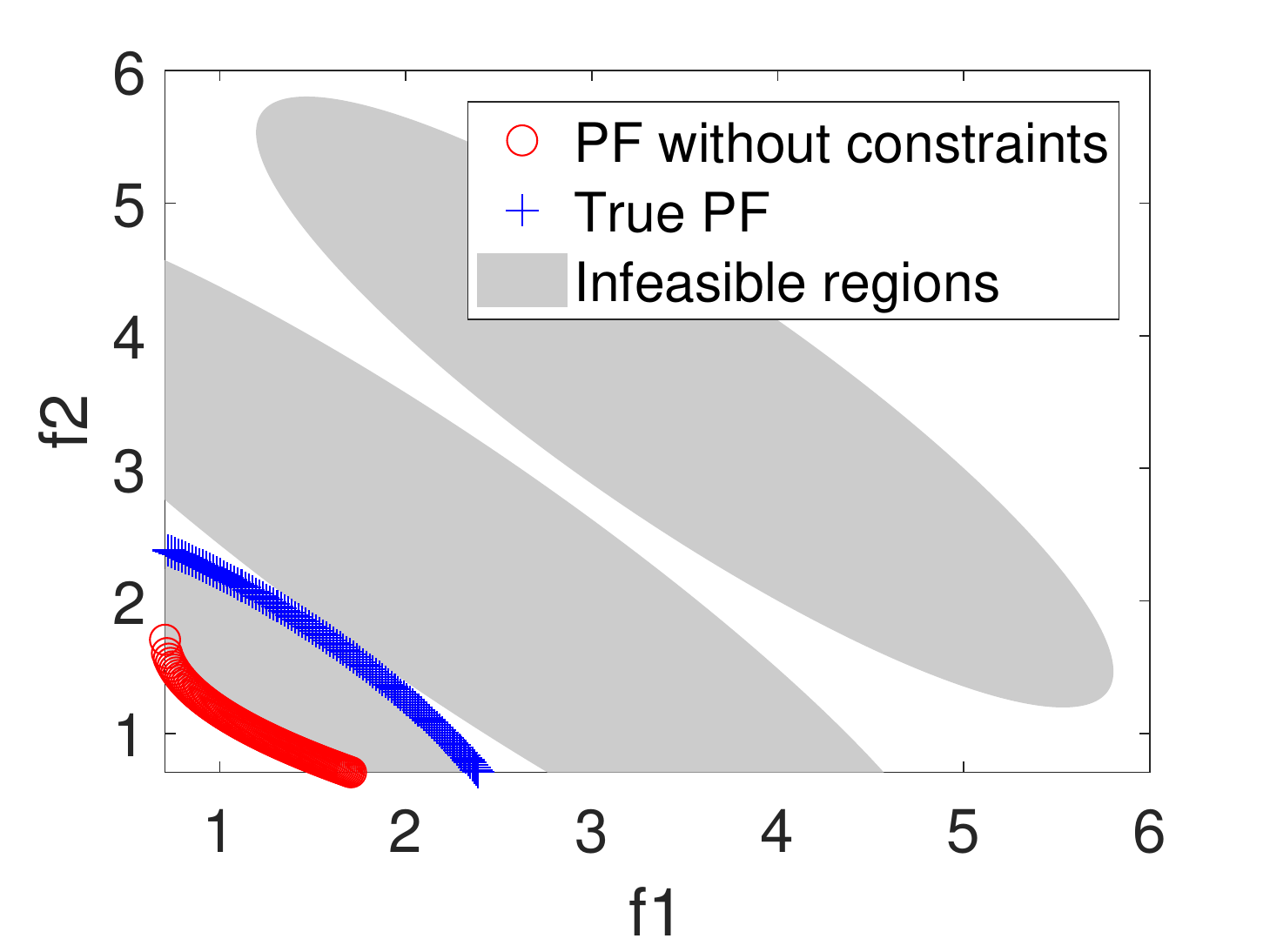}\\
			\centering{\scriptsize{(b) LIR-CMOP7}}
		\end{minipage}
		\hspace{0.5cm}
        \begin{minipage}[t]{0.28\linewidth}
			\includegraphics[height = 4.5cm]{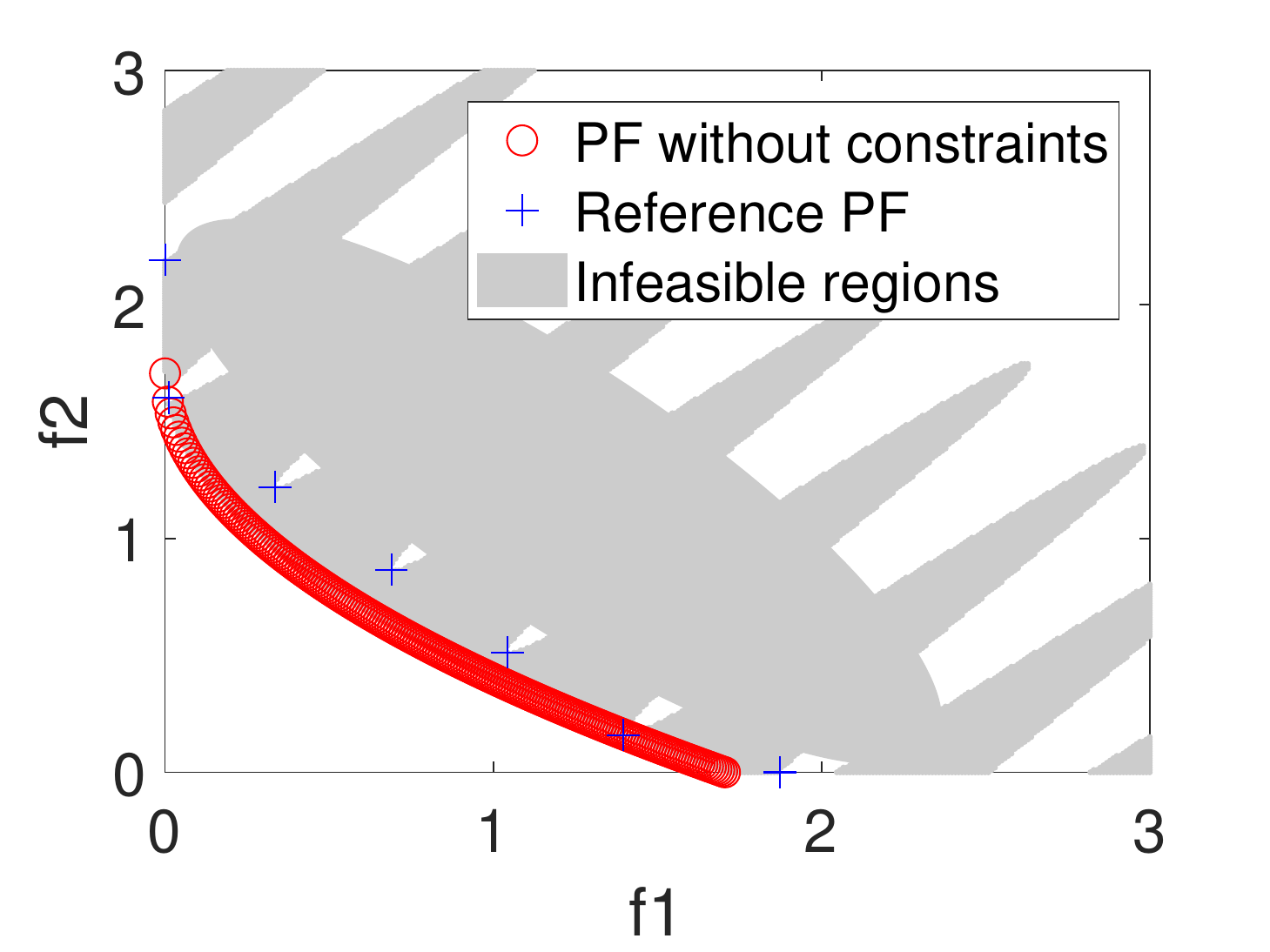}\\
			\centering{\scriptsize{(c) LIR-CMOP11}}
		\end{minipage}
		\hspace{0.5cm}
	\end{tabular}
\caption{\label{fig:pf}
Illustrations of the feasible and infeasible regions of LIR-CMOP2, LIR-CMOP7 and LIR-CMOP11, corresponding to three different types of difficulties as discussed in \cite{fan2019dascmop}}
\end{figure*}

In order to further discuss the advantages of the proposed PPS-M2M in solving CMOPs, we plot non-dominated solutions achieved by each algorithm on LIR-CMOP2, LIR-CMOP7 and LIR-CMOP11 with the median HV values. The feasible and infeasible regions of LIR-CMOP2, LIR-CMOP7 and LIR-CMOP11, corresponding to three different types of difficulties \cite{fan2019dascmop}, are plotted in Fig. \ref{fig:pf}.

\begin{figure*}
	\begin{tabular}{cc}
		\begin{minipage}[t]{0.25\linewidth}
			\includegraphics[width = 5cm]{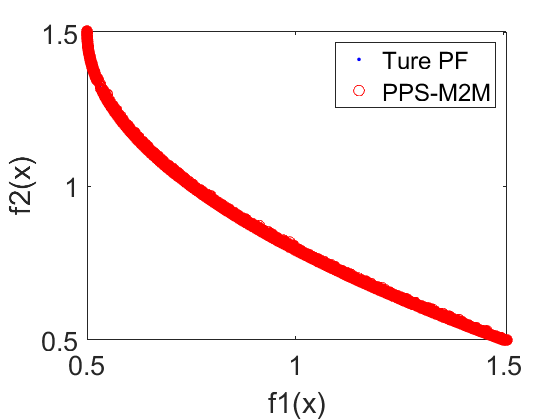}\\
			\centering{\scriptsize{(a) PPS-M2M}}
		\end{minipage}
        \begin{minipage}[t]{0.25\linewidth}
			\includegraphics[width = 5cm]{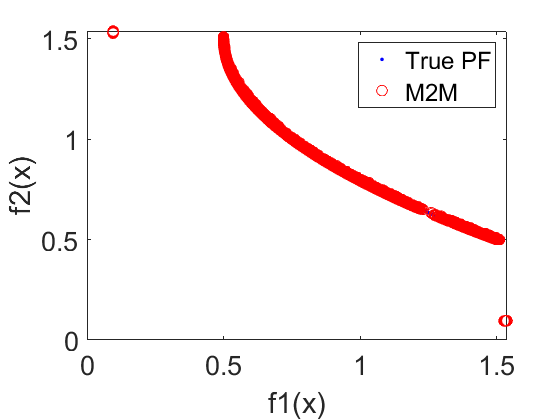}\\
			\centering{\scriptsize{(b) M2M}}
		\end{minipage}
        \begin{minipage}[t]{0.25\linewidth}
			\includegraphics[width = 5cm]{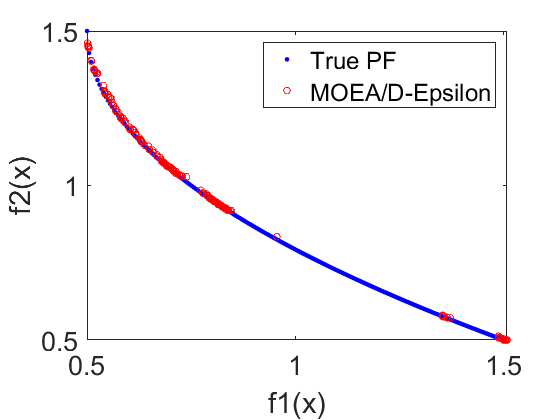}\\
			\centering{\scriptsize{(c) MOEA/D-Epsilon}}
		\end{minipage}
        \begin{minipage}[t]{0.25\linewidth}
			\includegraphics[width = 5cm]{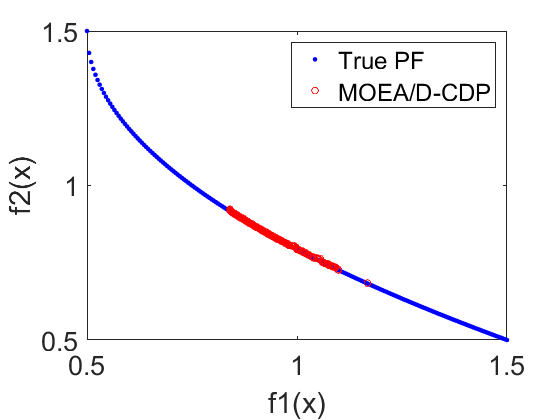}\\
			\centering{\scriptsize{(d) MOEA/D-CDP}}
		\end{minipage}
	\end{tabular}
	
    \begin{tabular}{cc}	
        \begin{minipage}[t]{0.25\linewidth}
			\includegraphics[width = 5cm]{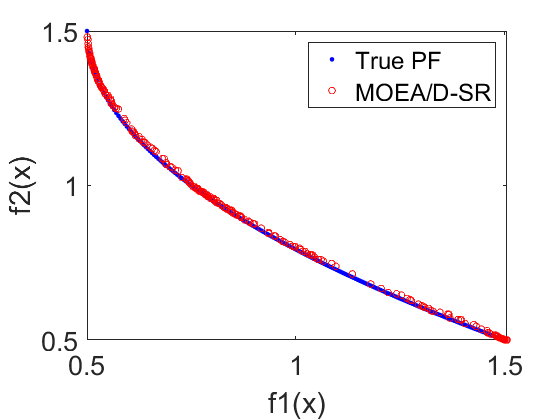}\\
			\centering{\scriptsize{(e) MOEA/D-SR}}
		\end{minipage}
		\begin{minipage}[t]{0.25\linewidth}
			\includegraphics[width = 5cm]{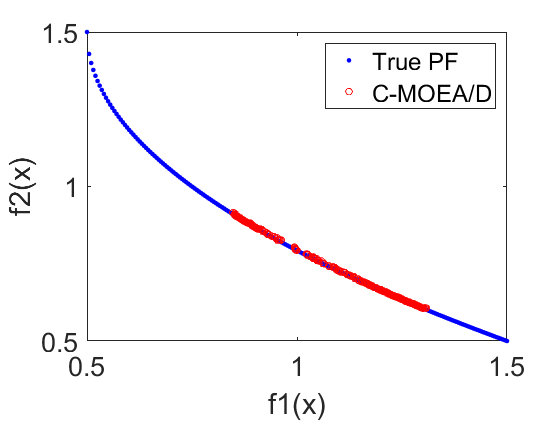}\\
			\centering{\scriptsize{(f) C-MOEA/D}}
		\end{minipage}
		\begin{minipage}[t]{0.25\linewidth}
			\includegraphics[width = 5cm]{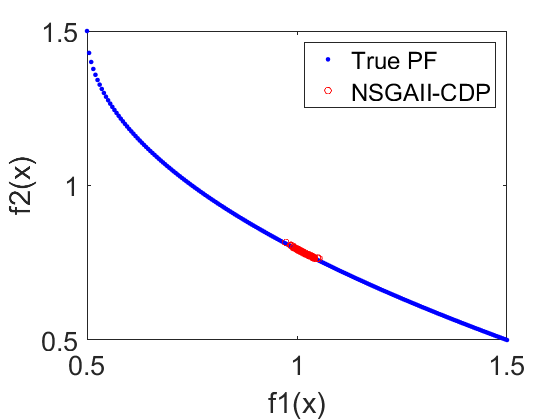}\\
			\centering{\scriptsize{(g) NSGA-II-CDP}}
		\end{minipage}
	\end{tabular}
	\caption{\label{fig:LIR-CMOP2-pops} The non-dominated solutions achieved by each algorithm on LIR-CMOP2 with the median HV values.}
\end{figure*}

As shown in Fig. \ref{fig:pf}(a), the feasible region of LIR-CMOP2 is very small, which is a feasibility-hard problem. Non-dominated solutions achieved by each algorithm on LIR-CMOP2 with the median HV values are plotted in Fig. \ref{fig:LIR-CMOP2-pops}. We can observe that the proposed PPS-M2M can converge to the true PF and has good diversity, as shown in the Fig. \ref{fig:LIR-CMOP2-pops}(a). The convergence of M2M is worse than that of PPS-M2M, as shown in Fig. \ref{fig:LIR-CMOP2-pops}(b). For the rest of five algorithms, their diversity is worse than that of PPS-M2M, as shown in Fig. \ref{fig:LIR-CMOP2-pops}(c)-(g).

%

\begin{figure*}
	\begin{tabular}{cc}
		\begin{minipage}[t]{0.25\linewidth}
			\includegraphics[width = 5cm]{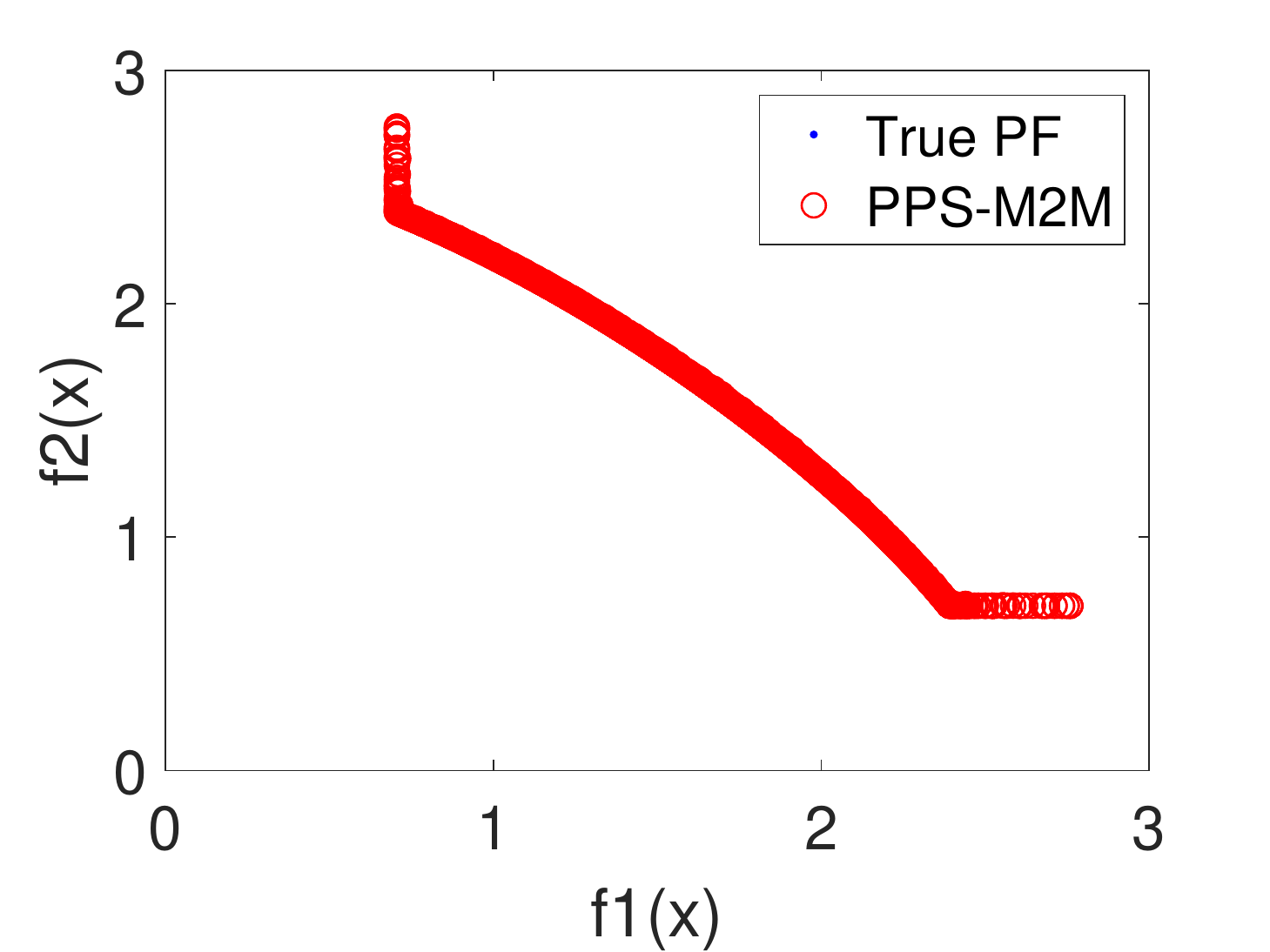}\\
			\centering{\scriptsize{(a) PPS-M2M}}
		\end{minipage}
        \begin{minipage}[t]{0.25\linewidth}
			\includegraphics[width = 5cm]{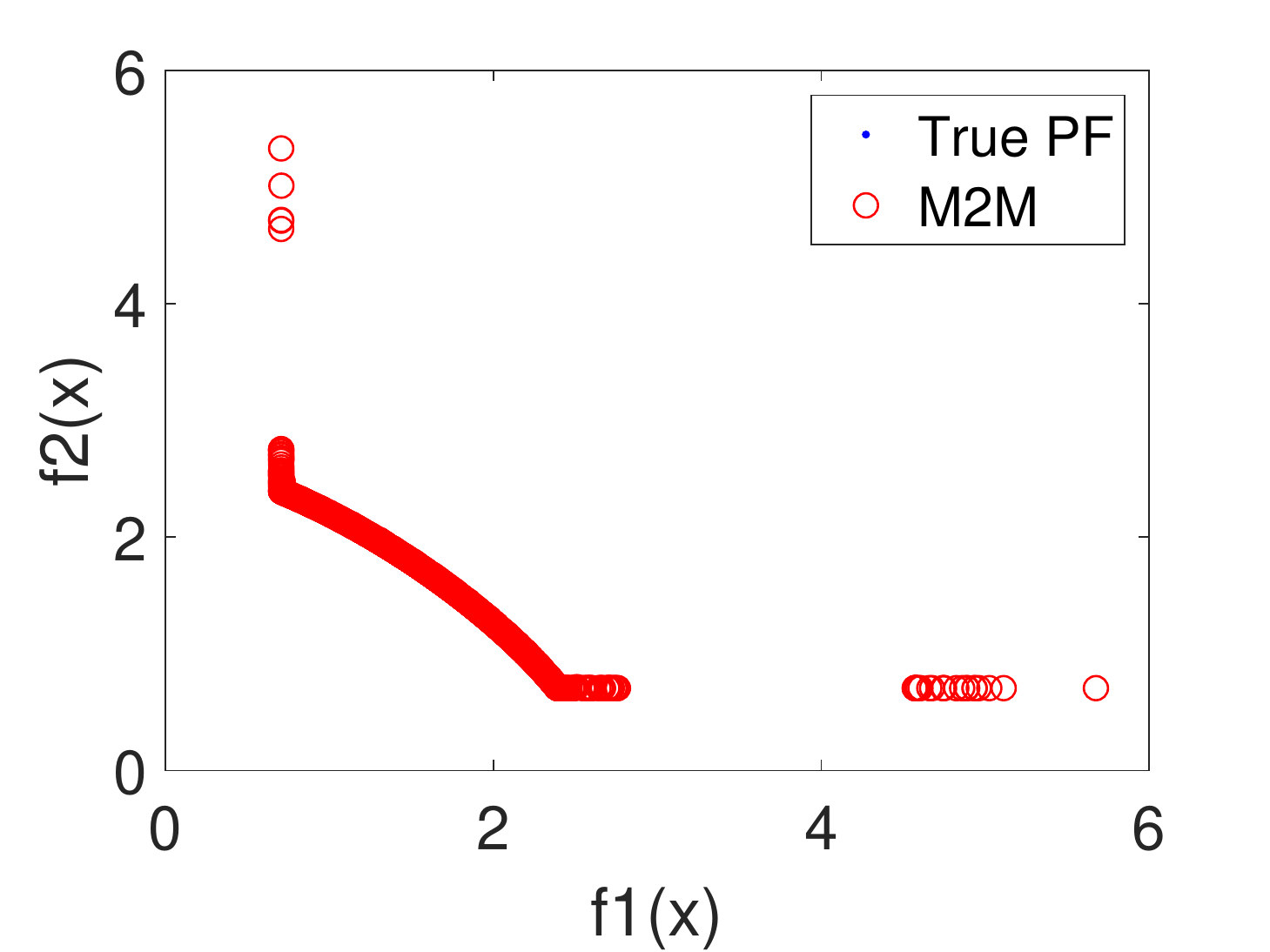}\\
			\centering{\scriptsize{(b) M2M}}
		\end{minipage}
        \begin{minipage}[t]{0.25\linewidth}
			\includegraphics[width = 5cm]{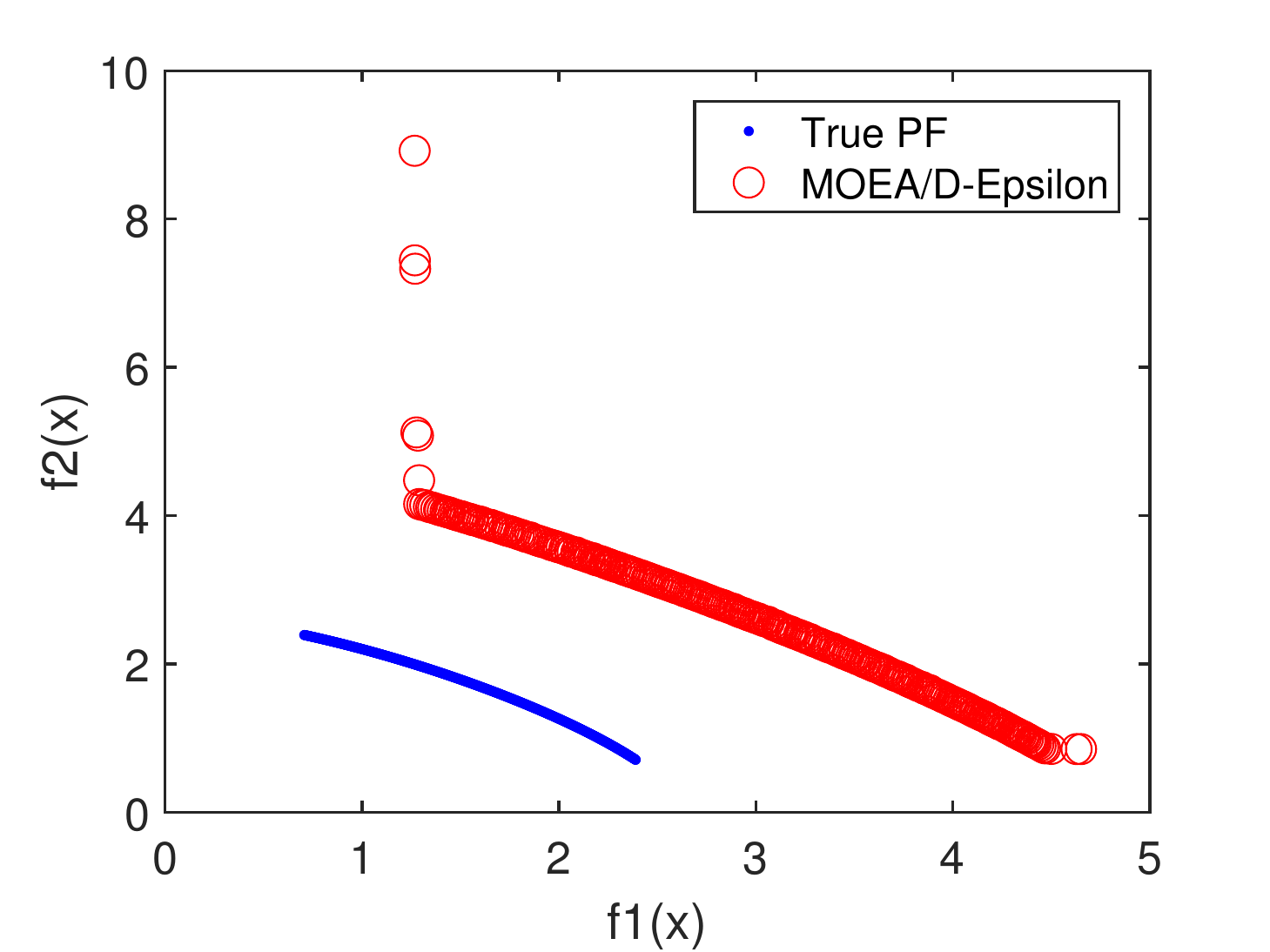}\\
			\centering{\scriptsize{(c) MOEA/D-Epsilon}}
		\end{minipage}
		\begin{minipage}[t]{0.25\linewidth}
			\includegraphics[width = 5cm]{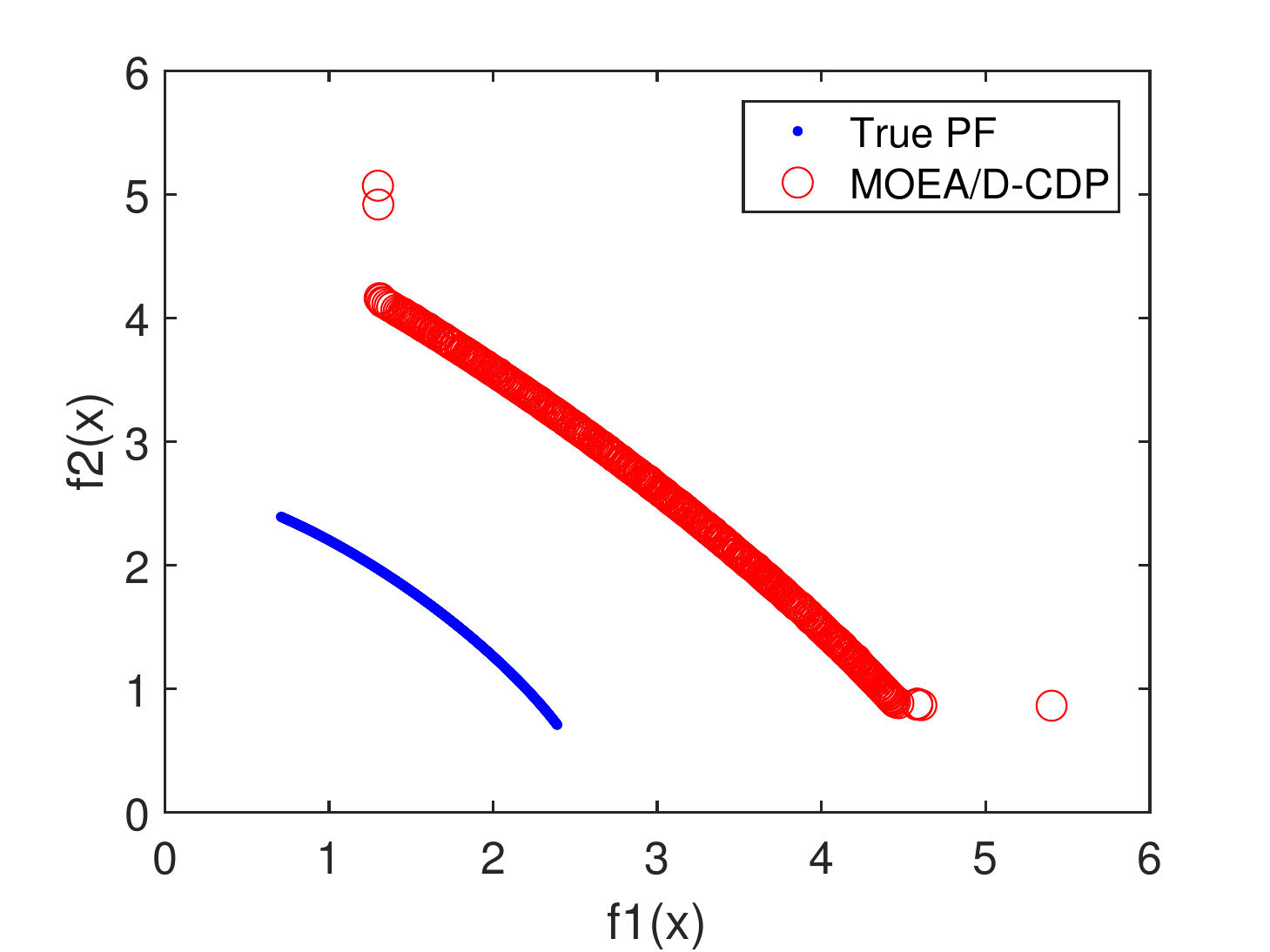}\\
			\centering{\scriptsize{(d) MOEA/D-CDP}}
		\end{minipage}
	\end{tabular}

    \vspace{0.2cm}
	\begin{tabular}{cc}
        \begin{minipage}[t]{0.25\linewidth}
			\includegraphics[width = 5cm]{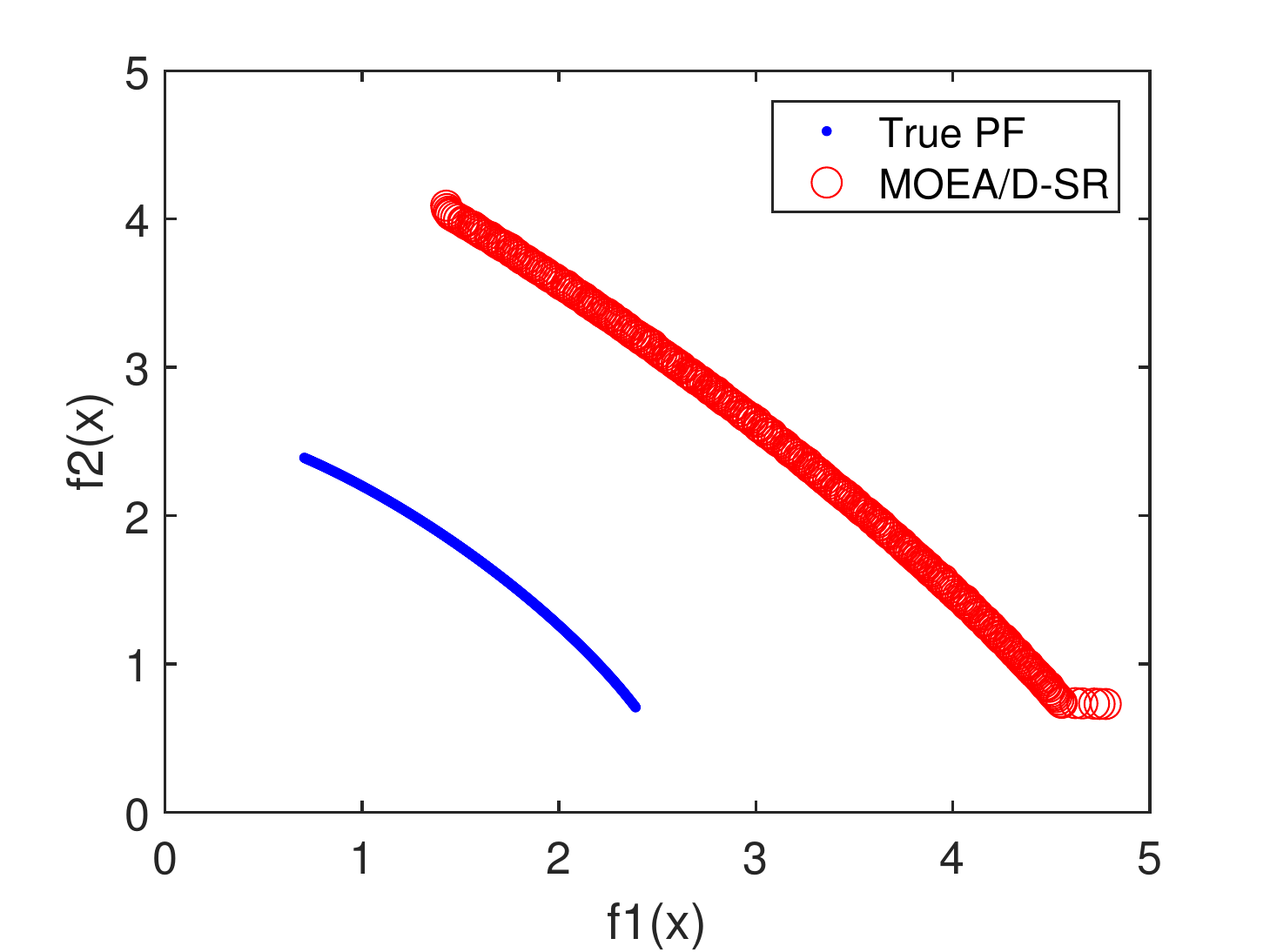}\\
			\centering{\scriptsize{(e) MOEA/D-SR}}
		\end{minipage}
        \begin{minipage}[t]{0.25\linewidth}
			\includegraphics[width = 5cm]{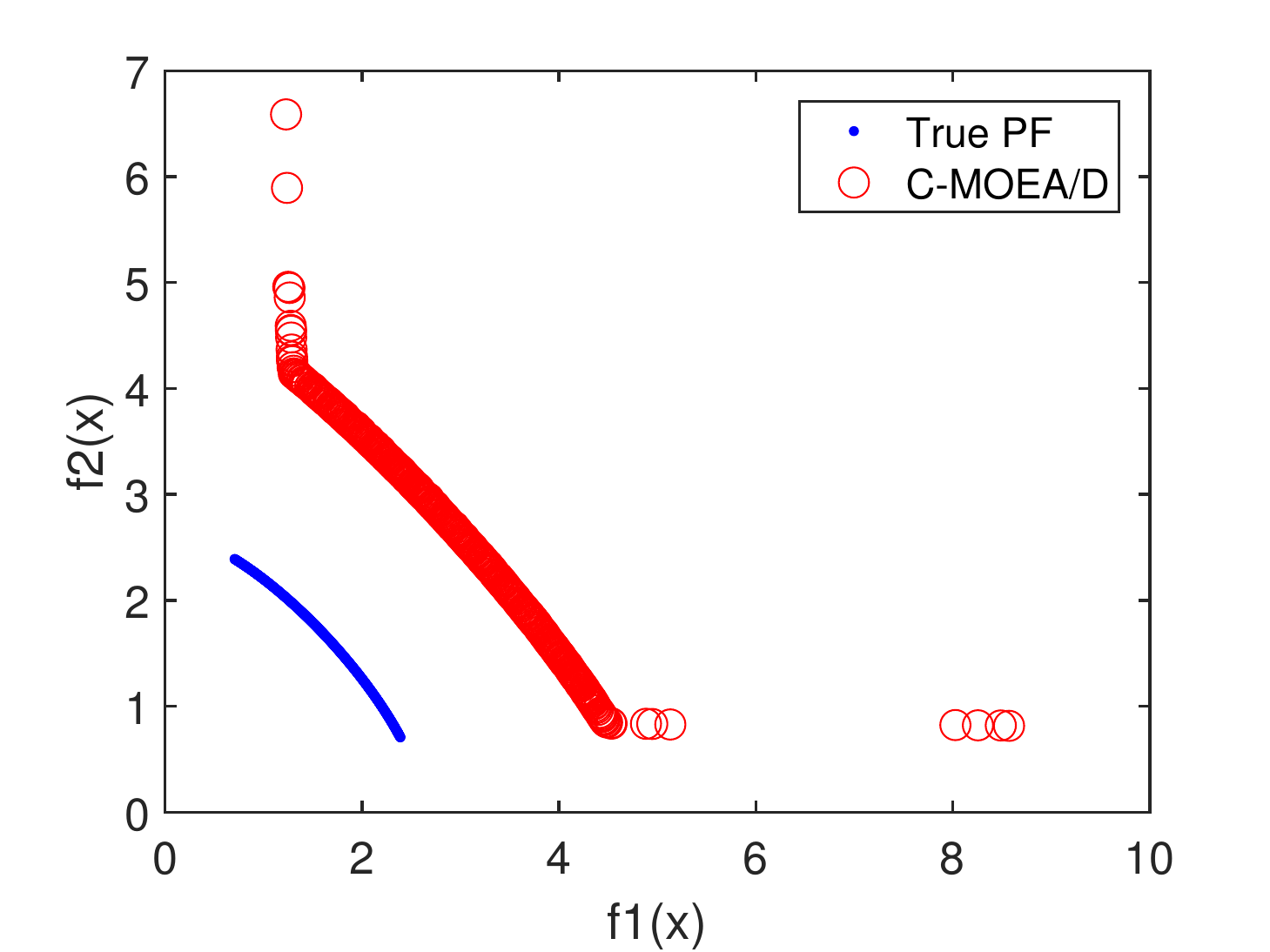}\\
			\centering{\scriptsize{(f) C-MOEA/D}}
		\end{minipage}
		\begin{minipage}[t]{0.25\linewidth}
			\includegraphics[width = 5cm]{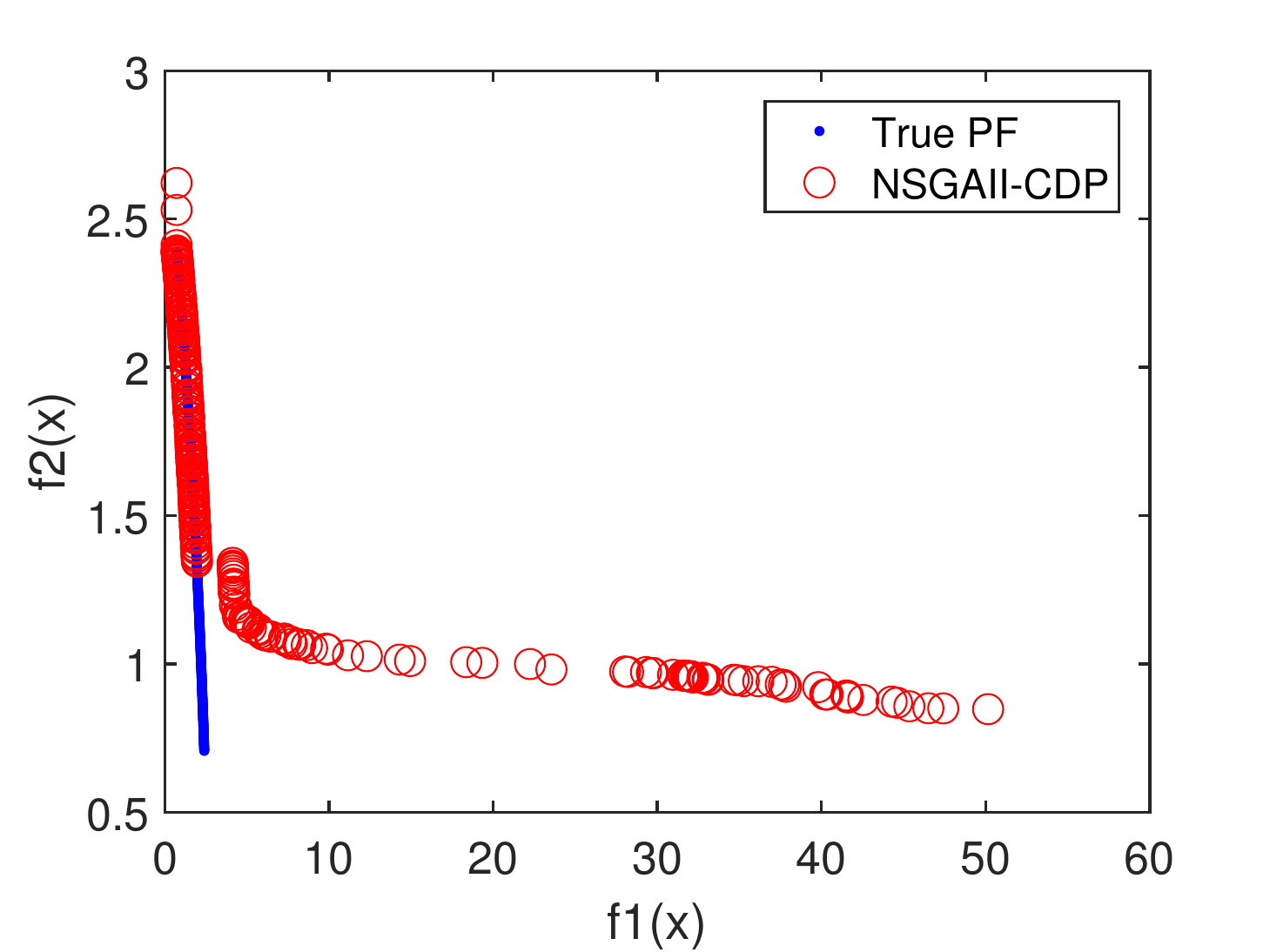}\\
			\centering{\scriptsize{(g) NSGA-II-CDP}}
		\end{minipage}
	\end{tabular}	
	\caption{\label{fig:LIR-CMOP7-pops} The non-dominated solutions achieved by each algorithm on LIR-CMOP7 with the median HV values.}
\end{figure*}

In LIR-CMOP7, there are three large non-overlapping infeasible regions in the front of the unconstrained PF, as illustrated in Fig. \ref{fig:pf}(b). In addition, the unconstrained PF is covered by one of the infeasible regions, which is a convergence-hard problem. The results of PPS-M2M and the other six algorithms (M2M, MOEA/D-Epsilon, MOEA/D-SR, MOEA/D-CDP, C-MOEA/D and NSGA-II-CDP) on LIR-CMOP7 are shown in Fig. \ref{fig:LIR-CMOP7-pops}. We can see that only PPS-M2M and M2M can get across the infeasible regions to reach the true PF, as illustrated in Fig. \ref{fig:LIR-CMOP7-pops}(a)-(b), while the other five algorithms (MOEA/D-Epsilon, MOEA/D-SR, MOEA/D-CDP, C-MOEA/D and NSGA-II-CDP) cannot converge to the true PF, as illustrated by Fig. \ref{fig:LIR-CMOP7-pops}(c)-(g). A possible reason is that two large infeasible regions in front of the PF hinder the way of populations of the five algorithms towards to the constrained PF.

\begin{figure*}
	\begin{tabular}{cc}
		\begin{minipage}[t]{0.25\linewidth}
			\includegraphics[width = 5cm]{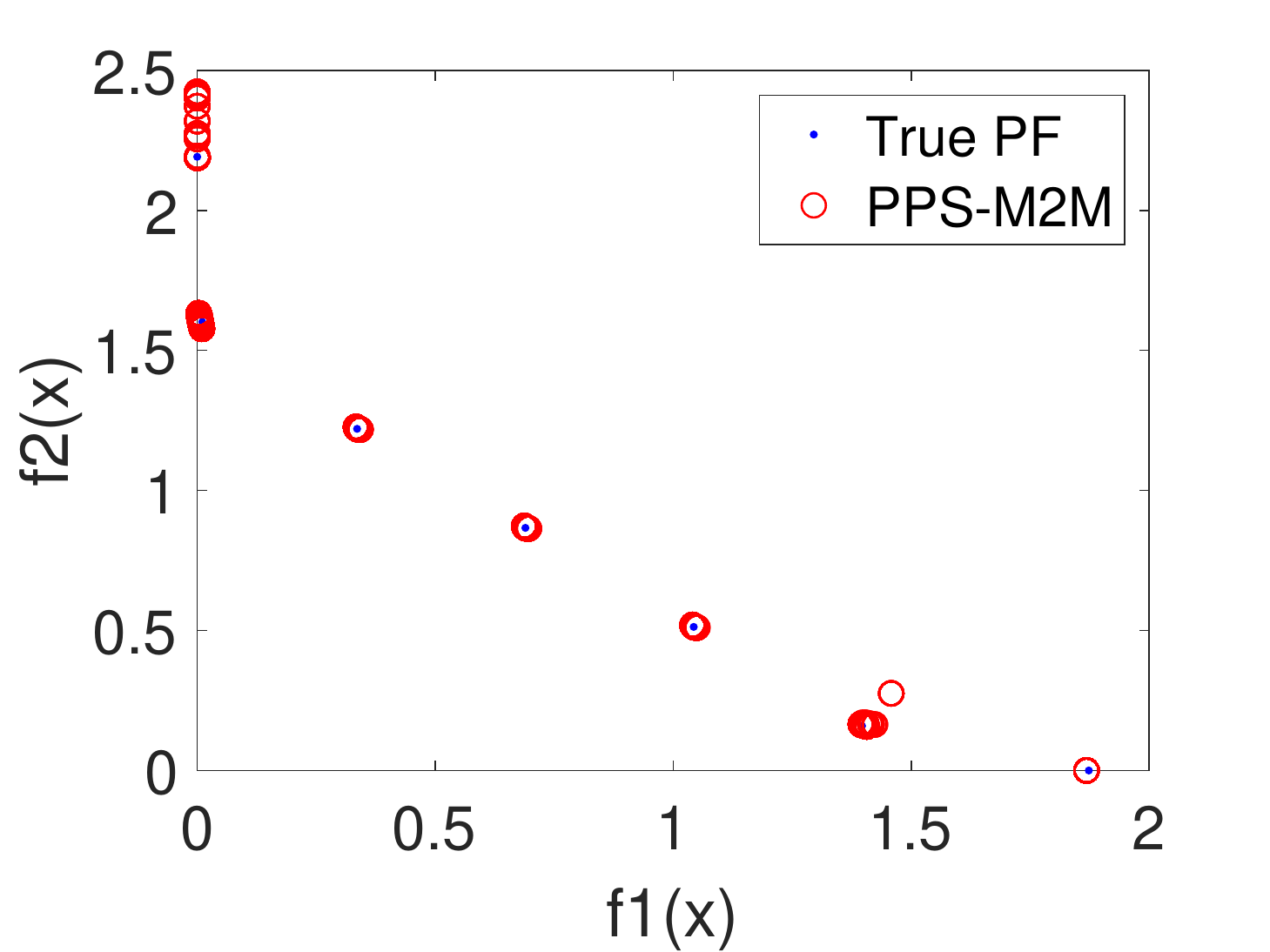}\\
			\centering{\scriptsize{(a) PPS-M2M}}
		\end{minipage}
        \begin{minipage}[t]{0.25\linewidth}
			\includegraphics[width = 5cm]{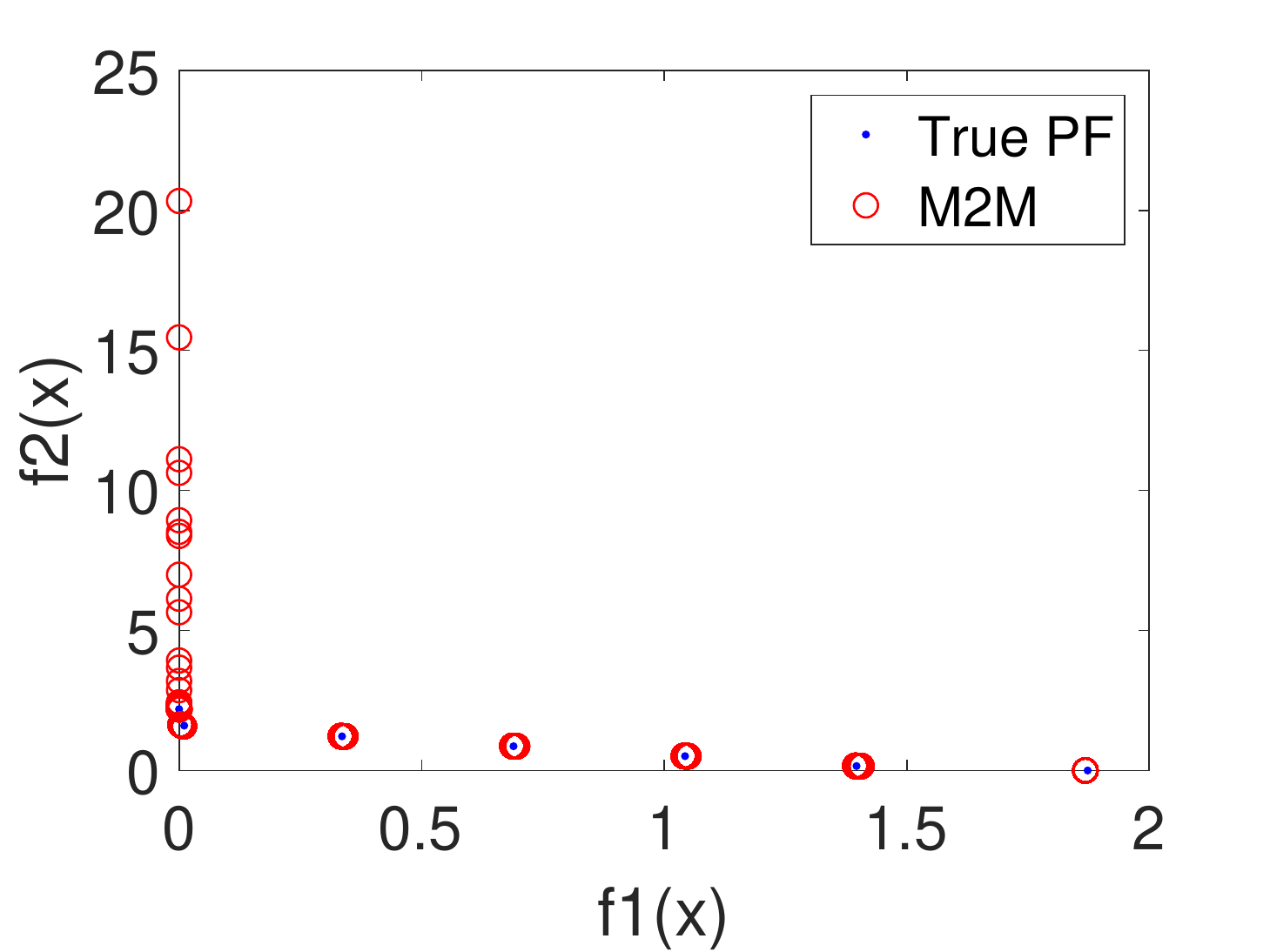}\\
			\centering{\scriptsize{(b) M2M}}
		\end{minipage}
        \begin{minipage}[t]{0.25\linewidth}
			\includegraphics[width = 5cm]{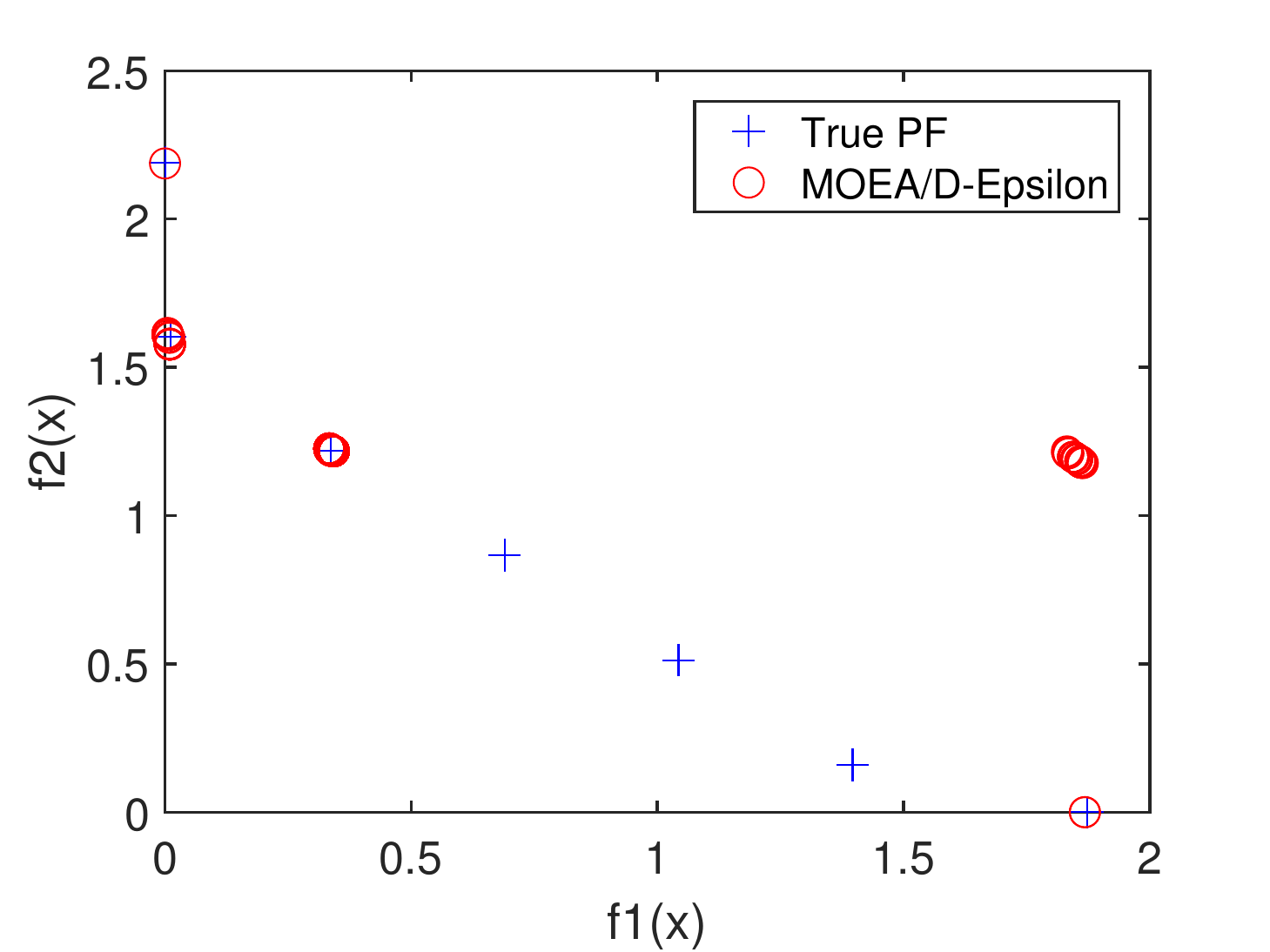}\\
			\centering{\scriptsize{(c) MOEA/D-Epsilon}}
		\end{minipage}
        \begin{minipage}[t]{0.25\linewidth}
			\includegraphics[width = 5cm]{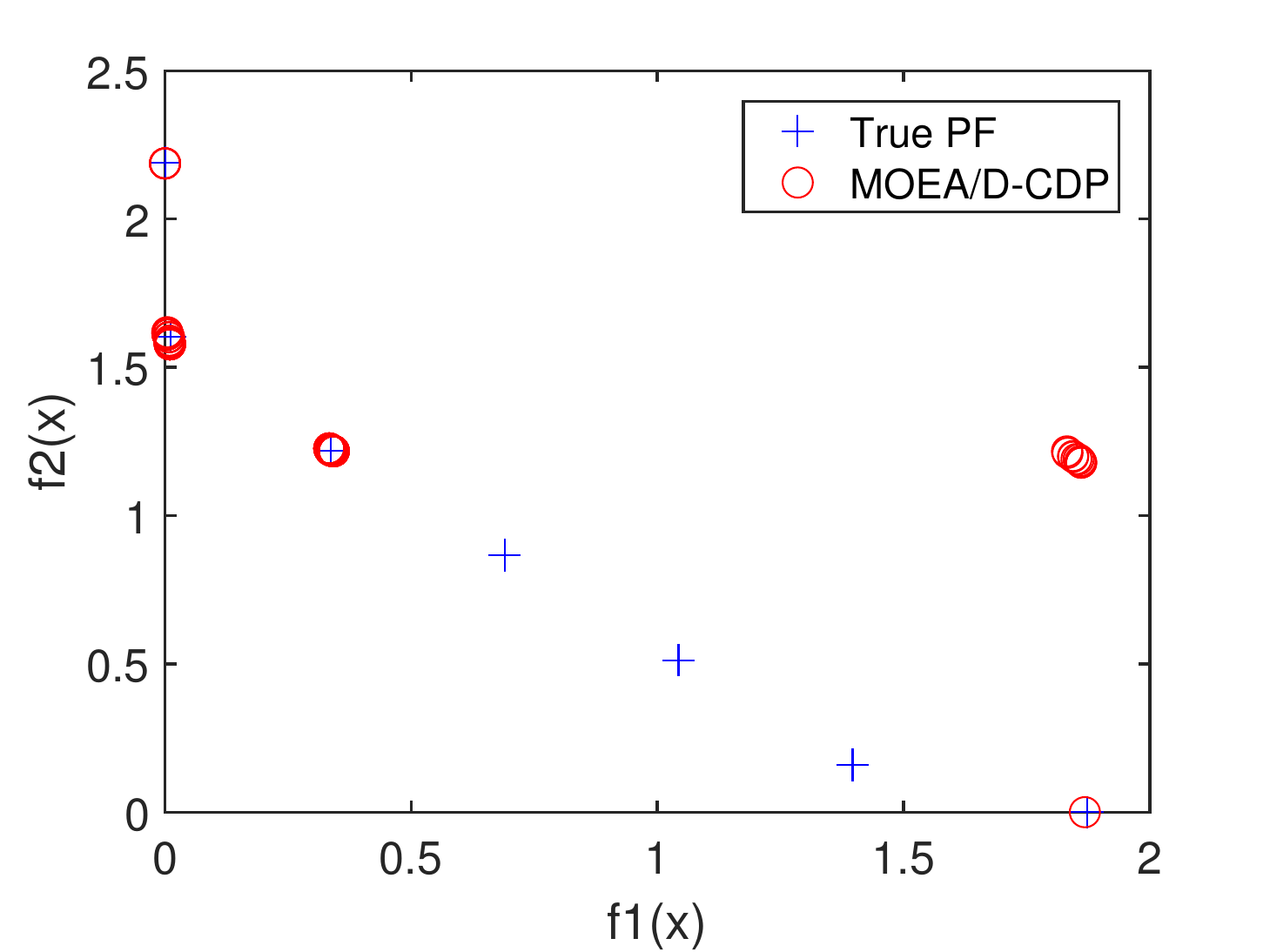}\\
			\centering{\scriptsize{(d) MOEA/D-CDP}}
		\end{minipage}
	\end{tabular}
	
    \begin{tabular}{cc}	
        \begin{minipage}[t]{0.25\linewidth}
			\includegraphics[width = 5cm]{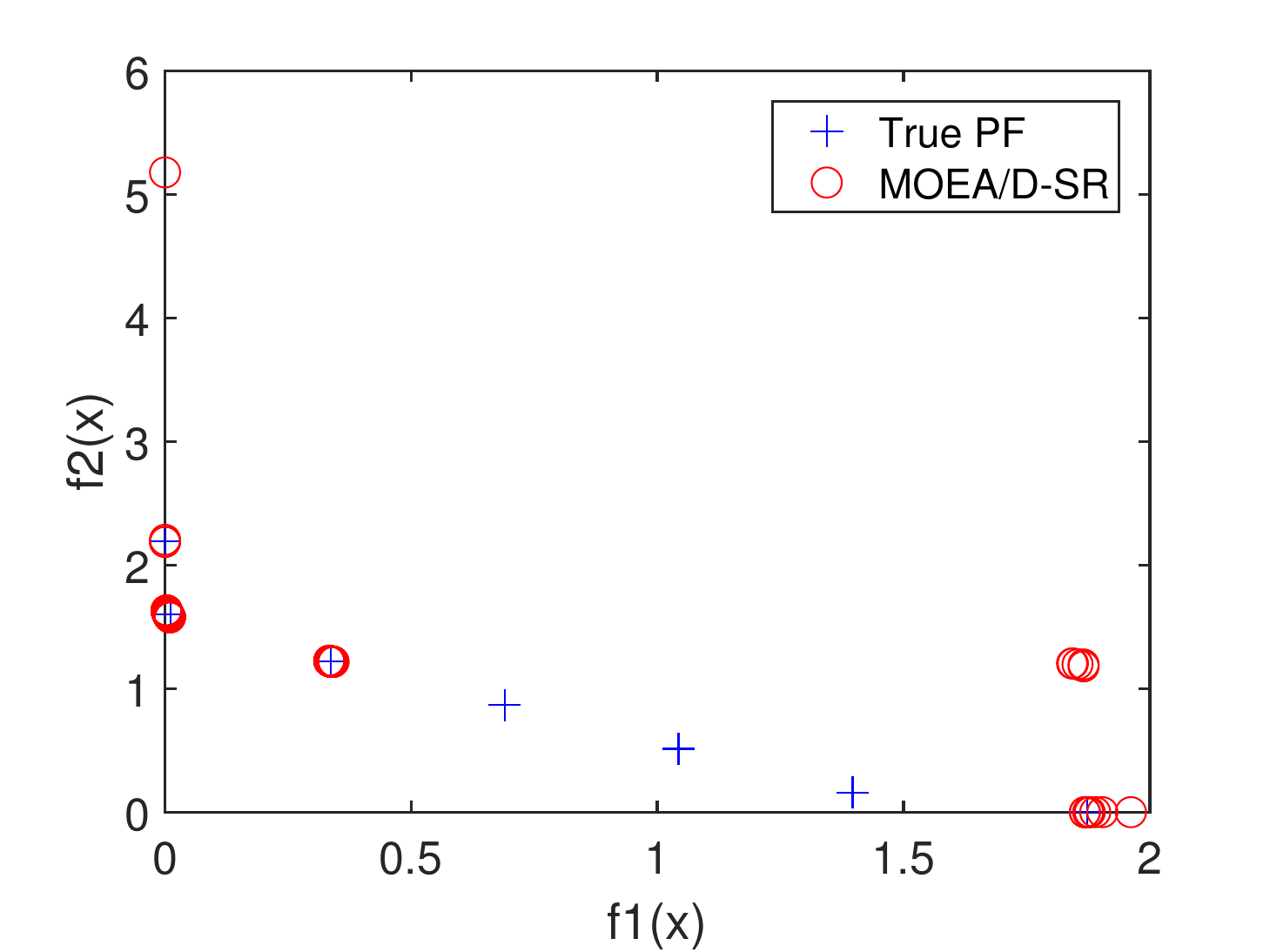}\\
			\centering{\scriptsize{(e) MOEA/D-SR}}
		\end{minipage}
		\begin{minipage}[t]{0.25\linewidth}
			\includegraphics[width = 5cm]{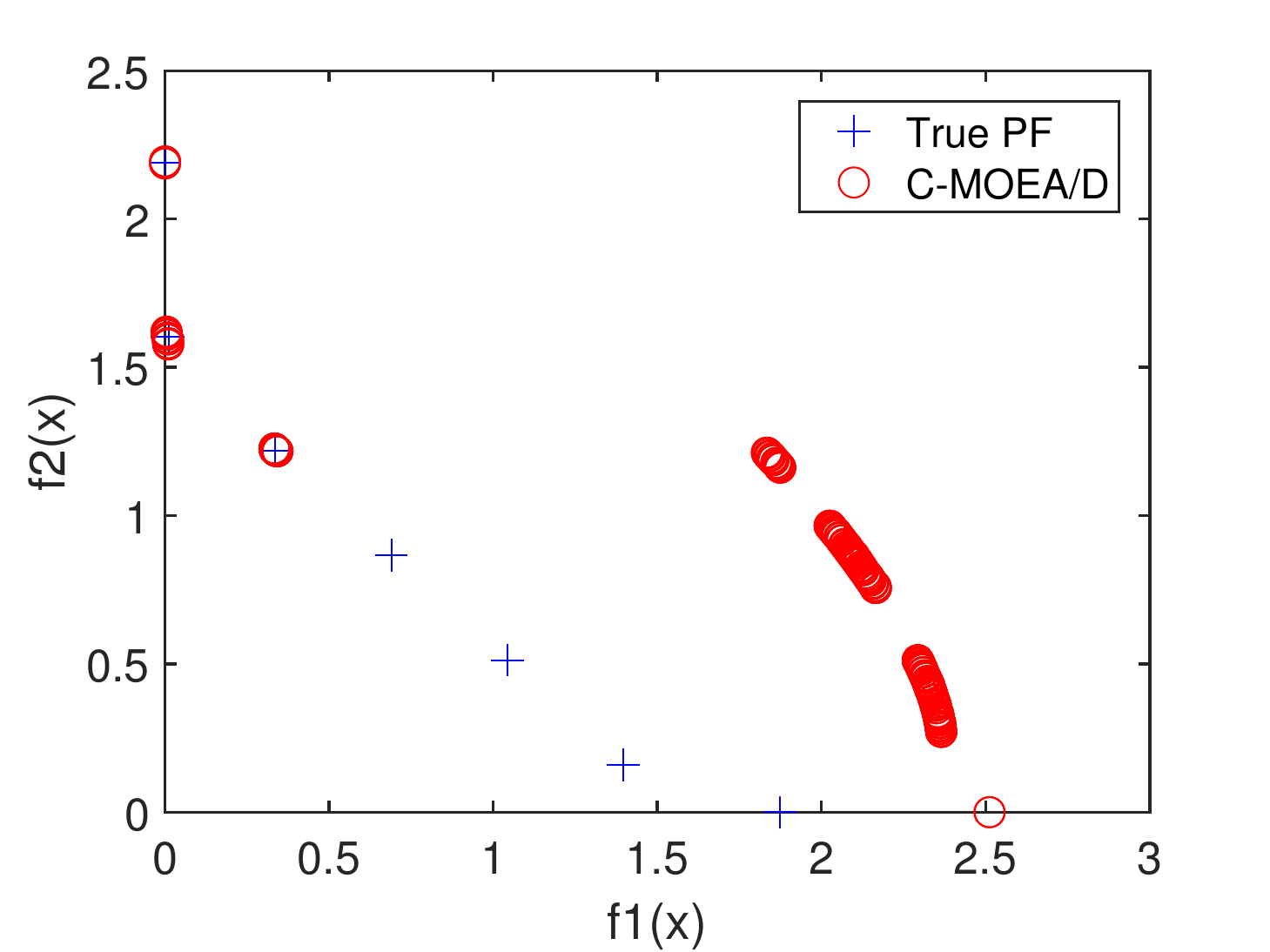}\\
			\centering{\scriptsize{(f) C-MOEA/D}}
		\end{minipage}
		\begin{minipage}[t]{0.25\linewidth}
			\includegraphics[width = 5cm]{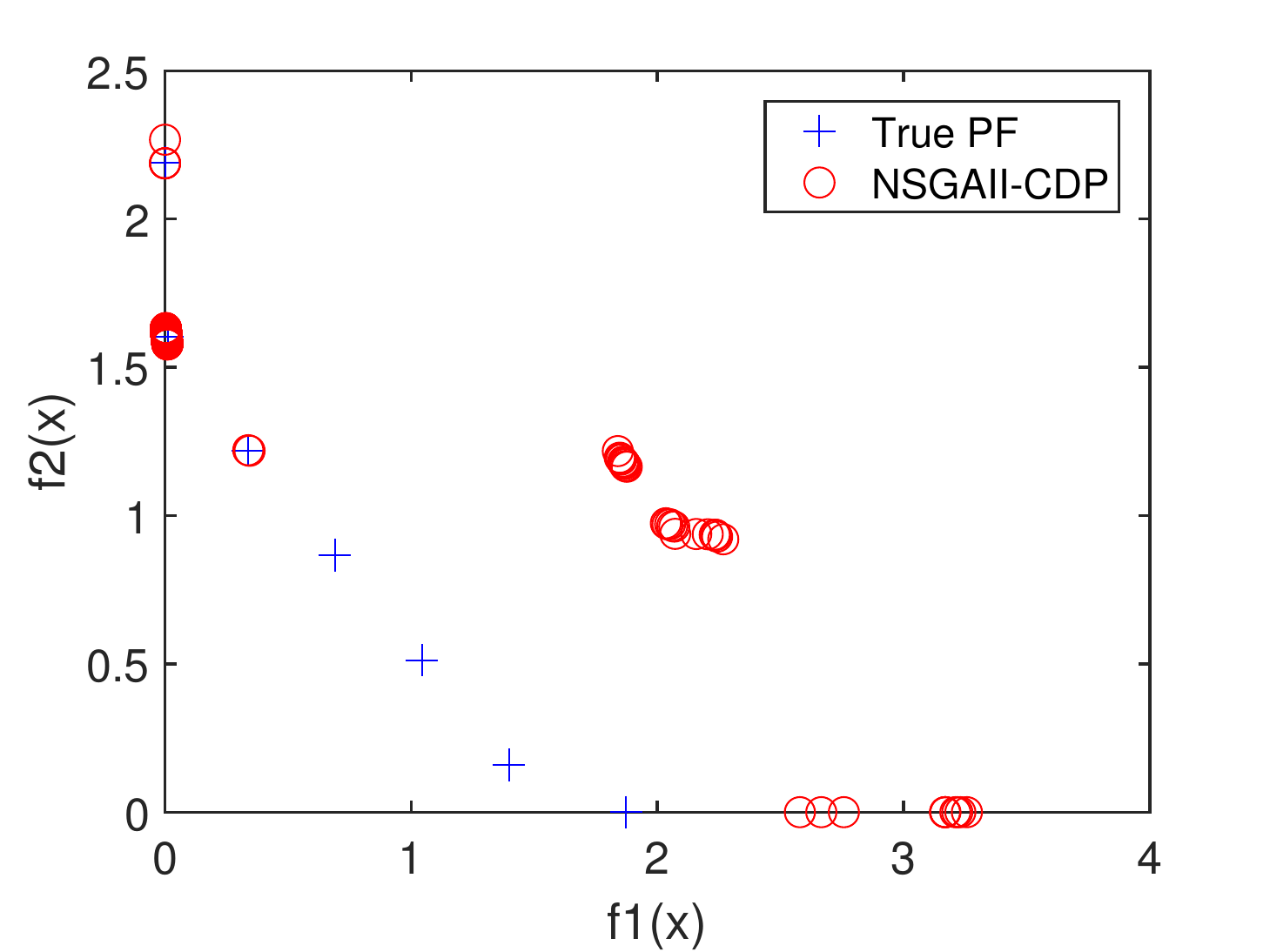}\\
			\centering{\scriptsize{(g) NSGA-II-CDP}}
		\end{minipage}
	\end{tabular}
	\caption{\label{fig:LIR-CMOP11-pops} The non-dominated solutions achieved by each algorithm on LIR-CMOP11 with the median HV values.}
\end{figure*}

The PF of LIR-CMOP11 is discrete as shown in Fig. \ref{fig:pf}(c), which is a problem with diversity-hardness. There are seven Pareto optimal solutions. Two solutions are located on the unconstrained PF, and five solutions are located on the constrained PF. Fig. \ref{fig:LIR-CMOP11-pops} shows the results of the seven tested algorithms on LIR-CMOP11. From Fig. \ref{fig:LIR-CMOP11-pops}(a)-(b), we can see that only PPS-M2M and M2M can find all the Pareto optimal solutions. Furthermore, the PPS-M2M has good convergence compared to the M2M. A possible reason is that each sup-population are combined into a whole population which is evolved by employing the improved epsilon constraint-handling method and the $\varepsilon$-dominance technique at the last ten percentages of the maximum generation. The other five algorithms (MOEA/D-Epsilon, MOEA/D-SR, MOEA/D-CDP, C-MOEA/D and NSGA-II-CDP) find only a part of Pareto optimal solutions, because infeasible regions block the way of populations of the other five algorithms towards to the true PF.


Based on the above observations and analysis, we can conclude that the proposed PPS-M2M outperforms the other six CMOEAs on most of the test instances.

\section{Conclusion}
\label{sec:conc}

This paper proposes a new algorithm (PPS-M2M) that combines a multi-objective to multi-objective (M2M) decomposition approach with a push and pull search (PPS) framework to deal with CMOPs. To be more specific, the search process of PPS-M2M is divided into two stages—namely, push and pull search processes. At the push search stage, PPS-M2M uses the M2M decomposition method to decompose a CMOP into a set of simple CMOPs which correspond to a set of sub-populations. Each simple CMOP is solved in a collaborative manner without considering any constraints, which can help the sub-populations effortlessly get across infeasible regions. Furthermore, some constrained landscape information can be estimated during the push search stage, such as the ratio of feasible to infeasible solutions and the maximum overall constraint violation, which can be further employed to guide the parameter settings of constraint-handling mechanisms in the pull search stage. When the max rate of change between ideal and nadir points is less than or equal to a predefined threshold, the search process is switched to the pull search stage. At the beginning of the pull search stage, the infeasible solutions of each sup-population achieved in the push stage are pulled to the feasible and non-dominated regions by adopting an improved epsilon constraint-handling approach. At the last ten percentages of the maximum generation, each sup-population are combined into a whole population which is evolved by employing the improved epsilon constraint-handling method and the $\varepsilon$-dominance technique. The comprehensive experimental results demonstrate that the proposed PPS-M2M outperforms the other six CMOEAs on most of the benchmark problems significantly.

It is also worth noting that there are few studies regarding using the M2M decomposition method to solve CMOPs. In this paper, the proposed PPS-M2M provides a feasible method. Obviously, improving the performance of PPS-M2M requires a lot of work, such as, enhanced constraint-handling mechanisms in the pull search stage, and integrating machine learning methods to allocate computational resources dynamically into sub-populations of the PPS-M2M method. In addition, some other benchmark problems (e.g. CF1-CF10 \cite{zhang2008multiobjective}) and real-world optimization problems will be investigated by using the proposed PPS-M2M.

\section*{Acknowledgement}
This work was supported in part by the National Natural Science Foundation of China under Grant (61175073, 61300159, 61332002, 51375287) , the Guangdong Key Laboratory of Digital signal and Image Processing, the Science and Technology Planning Project of Guangdong Province (2013B011304002)and the Project of Educational Commission of Guangdong Province, China 2015KGJHZ014).



\section*{Reference}
 \bibliographystyle{elsarticle-num}
 \bibliography{pps_m2m.bib}





\end{document}